\documentclass[sigconf]{acmart}

\def\vec#1{\mathchoice{\mbox{\boldmath$\displaystyle#1$}}
{\mbox{\boldmath$\textstyle#1$}}
{\mbox{\boldmath$\scriptstyle#1$}}
{\mbox{\boldmath$\scriptscriptstyle#1$}}}
\usepackage{booktabs} % For formal tables
\usepackage{amsmath}
\usepackage{algorithm}
\usepackage[noend]{algpseudocode}
\usepackage[T1]{fontenc}
\usepackage{tabularx}
\usepackage{multirow}
\usepackage{array}
\usepackage{pbox}
\usepackage{balance}
\usepackage{color,soul}
\usepackage{nameref}
\pdfmapfile{+lmodern.map}
\usepackage{microtype}
\usepackage{pgfplots}
\usepackage{graphicx}
\usepackage{subcaption}
\usepackage{tikz}

\usepackage{url}
\usepackage{bm}

\newcommand{\A}{{\small\textcolor{darkgreen}{TRAD}}}
\newcommand{\B}{{\small\textcolor{blue}{PX}}}
\newcommand{\C}{{\small\textcolor{red}{VIGP}}}

\definecolor{darkgreen}{rgb}{0.30, 0.50, 0.0}

\makeatletter
\renewcommand{\ALG@name}{Pseudocode}
\makeatother
\newcolumntype{L}[1]{>{\raggedright\let\newline\\\arraybackslash\hspace{0pt}}m{#1}}
\newcolumntype{C}[1]{>{\centering\let\newline\\\arraybackslash\hspace{0pt}}m{#1}}
\newcolumntype{R}[1]{>{\raggedleft\let\newline\\\arraybackslash\hspace{0pt}}m{#1}}

\graphicspath{{images/}}

\copyrightyear{2025}
\acmYear{2025}
\setcopyright{cc}
\setcctype{by}
\acmConference[GECCO '25]{Genetic and Evolutionary Computation Conference}{July 14--18, 2025}{Malaga, Spain}
\acmBooktitle{Genetic and Evolutionary Computation Conference (GECCO '25), July 14--18, 2025, Malaga, Spain}\acmDOI{10.1145/3712256.3726426}
\acmISBN{979-8-4007-1465-8/2025/07}

\hypersetup{draft}

\hypersetup{colorlinks=true,urlcolor=blue}

\usepackage[normalem]{ulem}
\useunder{\uline}{\ulined}{}%
\DeclareUrlCommand{\bulurl}{}
\begin{document}
	\title{Subfunction Structure Matters: \\ A New Perspective on Local Optima Networks}

    \author{Sarah L. Thomson}
\affiliation{%
  \institution{Edinburgh Napier University}
  \city{Edinburgh}
  \country{Scotland, UK}}
  \email{s.thomson4@napier.ac.uk}

  \author{Michal W. Przewozniczek}
\affiliation{%
  \institution{Wroclaw University of Science and Technology}
  \city{Wroclaw}
  \country{Poland}}
  \email{michal.przewozniczek@pwr.edu.pl}
 	
	\renewcommand{\shortauthors}{}

    \begin{abstract}
Local optima networks (LONs) capture fitness landscape information. They are typically constructed in a black-box manner; information about the problem structure is not utilised. This also
applies to the \emph{analysis} of LONs: knowledge about the problem, such
as interaction between variables, is not considered. We challenge
this status-quo with an alternative approach: we consider how LON
analysis can be improved by incorporating subfunction-based information — this can either be known a-priori
or learned during search. To this end, LONs are constructed
for several benchmark pseudo-boolean problems using three approaches: firstly, the standard algorithm; a second algorithm which
uses deterministic grey-box crossover; and a third algorithm
which selects perturbations based on learned information about
variable interactions. Metrics related to subfunction changes in
a LON are proposed and compared with metrics from
previous literature which capture other aspects of a LON. Incorporating problem structure in LON construction and analysing it can
bring enriched insight into optimisation dynamics.
Such information may be crucial to understanding the difficulty of solving a given problem with state-of-the-art linkage learning
optimisers. In light of the results, we suggest incorporation of problem structure as an alternative paradigm in landscape analysis for
problems with known or suspected subfunction structure.
    \end{abstract}
	
	

	
	\maketitle
	
	\section{Introduction}

In recent years fitness landscape analysis has increased in popularity \cite{malan2021survey}, probably because of expanding interest in the notion of explainability \cite{zhou2024evolutionary} for optimisation algorithms. Local optima networks (LONs) \cite{ochoa2008study} are a landscape analysis tool which have been successfully used to help explain optimisation dynamics across several domains \cite{verel2012local,whitley2022local,treimun2020modelling,teixeira2022understanding}. Typically, LONs are visualised and metrics are computed to achieve these aims. Many metrics have been proposed: complex network measurements such as node degrees, clustering coefficients, and assortativity \cite{ochoa2008study}, metrics related to neutrality and plateaus at the level of local optima \cite{mostert2019insights}, to fractal dimension \cite{thomson2022fractal}, and to the notion of \emph{funnels} \cite{ochoa2017understanding} --- among others. Something which has not been incorporated in the analysis of LONs is knowledge of problem structure: in particular, the existence of subfunctions. 
Many optimisation problems can be represented in a subfunction-based manner. For instance, the $k$-bounded problems \cite{transTokBounded} can be represented as a sum of subfunctions which take no more than \textit{k} arguments. The subfunction structure can be known in advance \cite{decFunc,cyclicTrap} or learned during search \cite{ellGomea}.
Previous works have constructed LONs for $k$-bounded problems such as MAX3SAT \cite{ochoa2019local} and NK landscapes \cite{chicano2017optimizing}; however, the measurements taken from the LONs did not consider subfunction structure which may be crucial to understand the effectiveness (or its lack) of state-of-the-art gray-box mechanisms \cite{pxForBinary,ilsDLED} and optimizers using linkage learning \cite{dgga,FIHCwLL}. Our main contribution in this paper is to address this gap. We construct LONs for well-known $k$-bounded problems and analyse them in a new way. Novel visualisations and metrics are proposed and used to compare algorithms and problem types. We find that incorporating subfunction structure into LON analysis is promising for bringing insight into optimisation dynamics on $k$-bounded problems. 

\section{Preliminaries}
    \subsection{Optimisation problems}
    \label{sec:back:optProblems}
All problems used in this study have objective functions which are subject to \emph{maximisation}. We now describe each of the problems in turn. 

\paragraph{Deceptive functions}
We consider problems whose solutions are represented by binary vectors of size $n$: $\vec{x}=(x_1,x_2,\ldots,x_{n})$. Deceptive functions \cite{decFunc} were proposed to support tools for modelling hard artificial problems for binary search spaces. They are frequently used as benchmarks in the research concerning Genetic Algorithms \cite{ltga,P3Original,FIHCwLL}. We consider standard and bimodal deceptive functions of unitation. The standard deceptive function of order $k$ is defined as:
\begin{equation}
\label{eq:trap}
    \mathit{trap}_k(u) = 
    \begin{cases}
    k - u - 1& ,u < k \\
    k & ,u = k\\
    \end{cases}
\end{equation}
where $u$ is the sum of binary values (so-called \textit{unitation}) of the function's argument. The standard deceptive trap function has one local optimum (for $u=0$) and one global optimum (for $u=k$). It may be considered hard to solve because starting from the random solution, any local search algorithm that greedily flips a single bit will converge to the suboptimal solution with $u=0$. Such a local search will find the optimal solution only if it is randomly hit or starts from the solution whose unitation is $u=k-1$, and the first flipped bit is the only '0' in such a genotype.\par

The bimodal deceptive function is defined as \cite{decBimodalOld}
\begin{equation}
\small
\label{eq:bimodal}
\mathit{bimTrap}_k(u) = 
\begin{cases}
k / 2 - |u - k/2| - 1 & ,u \neq k \land u \neq 0\\
k / 2 & ,u = k \lor u = 0\\
\end{cases}
\end{equation}

Bimodal deceptive functions have two optimal solutions (all '0's and all '1's) and $\binom{k/2}{k}$ or $\binom{k/2+1}{k}$ of locally optimal solutions for even and odd $k$, respectively.\par

\paragraph{Problem structure} Deceptive functions can be assembled to create larger problems, e.g., they can be concatenated. In such a problem, subfunctions do not share any arguments, and each subfunction can be optimised separately. However, they can form more sophisticated problems if they share variables. Then, the subfunctions will \textit{overlap}. The example of overlapping problems are cycling traps \cite{cyclicTrap}, where each deceptive function shares $o$ variables with its left and right neighbours. In \textit{conforming overlapping} problems, the optimal solution of each subfunction is a part of the globally optimal solution. In \textit{conflicting overlapping} problems, the optimal solution of each subfunction is not a part of the globally optimal solution. The differences between conflicting and conforming overlapping problems are discussed in subsequent sections. More information can be found in \cite{rdg3,overlapsKommar}.

\paragraph{NK Landscapes} were designed to model heavily overlapping problems. In NK Landscapes, each variable forms the subfunction argument set with $k$ subsequent variables. Thus, every subfunction takes $k+1$ variables and each variable is an argument of $k+1$ subfunction.

\paragraph{MAX-SAT} The maximum satisfiability problem (MAXSAT) is a real-world problem in which we are to satisfy clauses that consist of logical variables that may overlap. Any MAXSAT can be reformulated to MAX3SAT, where each clause contains three variables \cite{transTokBounded}. We consider artificial MAX3SAT instances and the generator employed in \cite{P3Original}, using the typical clause-to-variable ratio of $cr = 4.27$.\par

\subsection{Local Optima Networks}

\paragraph{LON nodes.} We consider a solution to be a \emph{local optimum} $lo_i$ if its fitness according to a function $f$ is better than the fitnesses within its neighbourhood $N$. In practice, the neighbourhood is often sampled rather than fully enumerated --- which is the case in this work. Formally: \(\forall{ neigh \in SN(lo_i)}:\) $f(lo_i) > f(neigh)$ (assuming maximisation, as is the case for this study) where $SN(lo_i)$ is the \emph{sampled} neighbourhood and $neigh$ is a particular neighbour. In a LON, the nodes are a set of local optima according to the definition just defined. 

\paragraph{LON edges.} Two local optima $lo_i$ and $lo_j$ have a LON edge traced between them under the condition that $lo_j$ can be reached from applying random perturbation to $lo_i$ and then subsequently local search [this type of edge has been termed an \emph{escape} edges \cite{verel2012local} in previous LON literature], and additionally $f(lo_j) \geq f(lo_i)$ (maximisation). This last criterion renders the edges {\em monotonic} in nature because they record only non-deteriorating, directed connections between local optima \cite{ochoa2017understanding}. Edges have weights: the frequency of transition between the two local optima. That is: $lo_j$ was reached through perturbation applied to $lo_i$ followed by local search. The set of edges is denoted by $E$.

\paragraph{Local optima network (LON)} We can now define a LON = $(LO,E)$, which comprises nodes $lo_i \in LO$ i.e. the local optima, and edges $e_{ij} \in E$ between pairs of nodes $lo_i$ and $lo_j$ with weight $w_{ij}$ iff $w_{ij} > 0$. 

\section{Related Work}
\label{sec:rw}

\subsection{Local optima networks}

The first ever study on LONs \cite{ochoa2008study} used NK landscapes as the domain, but did not consider subfunction structure in the analysis. The same is true for some subsequent studies on NK landscapes and LONs \cite{herrmann2016communities,thomson2017comparing}. Although LONs are usually constructed in a {\lq{black-box}\rq} manner \cite{verel2012local,ochoa2017understanding}; that is, with no problem structure information used --- there have been a few works which use grey-box optimisation to construct LONs. Chicano \emph{et al.} \cite{chicano2017optimizing} used greybox iterated local search with recombination to optimise high-dimensional NK landscapes and construct LONs. The LONs were visualised but not subject to subfunction analysis. Ochoa \emph{et al} \cite{ochoa2019local} also used greybox operators to construct LONs for MAX3SAT, but the metrics were {\lq{black-box}\rq} LON measurements such as the number of local optima plateaus. Two recent studies \cite{tari2023global,canonne2023combine} extract LONs using greybox operators for PUBOi and NKQ problems, respectively. Although subfunction structure is not considered in the LON analysis, the papers show a trend and interest in the community towards utilising problem structure in landscape analysis.

\subsection{Variable dependencies in evolutionary optimisation}
\label{sec:rw:varDeps}
Gray-box optimisation uses the \textit{a-priori} available knowledge about problem structure to leverage its effectiveness and efficiency \cite{whitleyNext}. This includes taking advantage of subfunction-based function representation that may utilise the \textit{additive form} defined as \cite{transTokBounded}:
    \begin{equation}
        \label{eq:additive}
        f(\vec{x})=\sum_{s=1}^{S} f_s(\vec{x}_{I_s}),
    \end{equation}
    where $I_s$ are subsets of $\{1,...,n\}$, which can overlap (i.e., do not have to be disjoint) and $S$ is the number of these subsets.\par

The additive form is convenient to represent the \textit{k}-bounded problems, i.e., the problems that can be represented by a sum of subfunctions where each subfunction does not take more than \textit{k} arguments \cite{transTokBounded}. Consider a situation in which the additive form of such a problem is known, which is the case in gray-box optimisation, and we have evaluated solution $\vec{x_a}$. Thus, we have computed the values of all subfunctions. By $\vec{x_a^g}$, we denote solution $\vec{x_a}$ with \textit{g}th gene flipped. To evaluate $\vec{x_a^g}$, we do not have to evaluate all subfunctions. Knowing the subfunction values for $\vec{x_a}$ is enough to compute only those subfunctions that take $x_g$ as an argument. This phenomenon is employed by \textit{partial evaluations} and decreases the computational costs of optimisation significantly \cite{GrayBoxWhitley,partialAnton}. Thus, the fewer subfunctions are modified by one move, the cheaper it is to evaluate.\par

In gray-box optimisation, it is typical to consider non-linear dependencies, i.e., variables $x_g$ and $x_h$ are dependent if \cite{linc}:
    \begin{equation}
    \small
    \label{eq:nonLinear}
        f(\vec{x}) + f(\vec{x}^{g,h}) \neq f(\vec{x}^g) + f(\vec{x}^h)
    \end{equation}
    where by $\vec{x}^{g}$, $\vec{x}^h$, and $\vec{x}^{g,h}$, are solutions obtained from $\vec{x}$ by modifying gene $g$, gene $h$ or both of them, respectively.\par

Frequently, the non-linear dependencies arise from decomposing the optimised function using the \textit{Walsh decomposition} \cite{heckendorn2002}

    \begin{equation}
    \small
    \label{eq:walsh-decomposition}
    f(\vec{x}) = \sum_{i=0}^{2^n-1} w_i \varphi_i(\vec{x}) 
    \end{equation}
    where $w_i \in \mathbb{R}$ is the $i$th Walsh coefficient, $\varphi_i(\mathbf{x}) =(-1)^{\mathbf{i}^\mathrm{T}\mathbf{x}}$ generates a sign, and $\mathbf{i} \in \{0,1\}^n$ is the binary representation of index $i$.  \par
If there exists at least one nonzero Walsh coefficient $w_i$ such that the $g$th and $h$th elements of $\mathbf{i}$ are equal to one, then variables $x_g$ and $x_h$ are non-linearly dependent \cite{ilsDLED}. In gray-box optimisation, the knowledge about variable dependencies is frequently represented by the Variable Interaction Graph (VIG) \cite{chicano2014efficient} that is a square matrix. VIG entry equals one if two variables are dependent and zero, otherwise. Gray-box operators frequently utilise this structure to obtain variation masks.

The Partition Crossover (PX) \cite{pxForBinary} is a recombination operator. It uses VIG to exchange only those variables that should be processed jointly. To mix two individuals $\vec{x_a}$ and $\vec{x_b}$, PX removes from VIG all dependencies for which at least one variable meets the condition $\vec{x_a}(i) = \vec{x_b}(i)$, where $\vec{x}(i)$ is the value of the \textit{i}th variable in $\vec{x}$. Then, PX joins variables in clusters using the remaining VIG dependencies. Each cluster can be used for mixing $\vec{x_a}$ and $\vec{x_b}$. Consider a PX offspring individual $\vec{x_a}'=\vec{x_a} + PX_{mask}(\vec{x_b})$ created by inserting genes from $\vec{x_b}$ marked by a PX mask into $\vec{x_a}$. The key feature of PX is that individuals $\vec{x_a}'$ and $\vec{x_b}'$ will refer only to those subfunction argument subsets that exist in $\vec{x_a}$ and $\vec{x_b}$ even if subfunctions overlap. If VIG concerns the non-linear dependencies, then it is guaranteed that $f(\vec{x_a}') + f(\vec{x_b}') = f(\vec{x_a}) + f(\vec{x_b})$. The above features make PX particularly useful in optimizing the overlapping problems \cite{dgga}.\par

VIG-based perturbation (VIGbp) \cite{ilsDLED} is a gray-box operator using VIG to produce variation masks. It randomly chooses a gene $x_r$ and considers all genes that are dependent on $x_r$ concerning VIG. If the number of genes dependent on $x_r$ is equal or lower than a user-defined parameter $\alpha$, then all genes are included by a mask. Otherwise, we randomly choose $\alpha$ genes from the $x_r$-dependent genes set. VIGbp is employed by the Iterated Local Search (ILS), which flips all genes marked by a mask and performs a local search on such perturbed genotypes. ILS aims to improve solutions step by step by improving only one part of a solution at a time. VIGbp perturbs a set of overlapping subfunctions instead of spreading the perturbation all over the genotype. VIGbp was shown to be highly effective in solving overlapping problems \cite{ilsDLED}.\par

\section{Problem structure and the expected LON}
\label{probStructAndLONs}

This section shows and analyses an example of an artificial problems and how LONs that describe its features should look. The objective is to present the key intuitions that will be helpful while considering the experimental results.\par

Consider $f_{sep}(x_0,x_1,...,x_8) = f_1(x_0,x_1,x_2) + f_2(x_3,x_4,x_5) + \\f_3(x_6,x_7,x_8)$, where each $f_s(\vec{x}_{I_s})=trap_3(u(\vec{x}_{I_s}))$. Subfunctions in $f_{sep}$ are separable, i.e., each of the argument sets (0,1,2), (3,4,5), and (6,7,8) can be optimised separately. If we start from a random solution, in most of the cases, we will converge to solution $000\text{ }000\text{ }000$. If we use \textbf{dependency-aware} ILS, e.g., ILS using VIGbp \cite{ilsDLED}, to optimise $f_{sep}$, then we will variate a variable subset that refers to one subfunction (a single $I_s$ in Formula \ref{eq:additive}). Thus, local optima that are worth showing in LON can be divided into four groups:

\begin{itemize}
    \item \textbf{Group 1} consists of a single solution: $000\text{ }000\text{ }000$.
    \item \textbf{Group 2} consists of three solutions: $111\text{ }000\text{ }000$, $000\text{ }111\text{ }000$, and $000\text{ }000\text{ }111$.
    \item \textbf{Group 3} consists of three solutions: $111\text{ }111\text{ }000$, $111\text{ }000\text{ }111$, and $000\text{ }111\text{ }111$.
    \item \textbf{Group 4} consists of a single optimal solution: $111\text{ }111\text{ }111$.    
\end{itemize}

\begin{figure}[h]
    \centering
    \includegraphics[width=0.7 \linewidth]{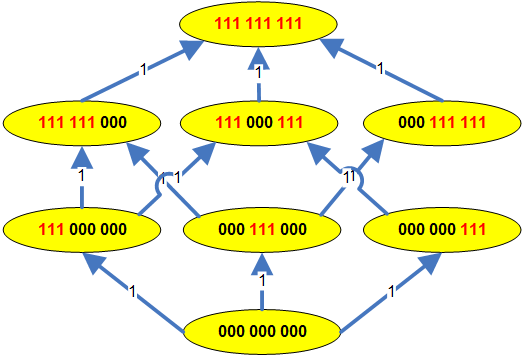}
    \caption{Decomposition-aware LON showing the number of modified subfunctions in each move}
    \label{fig:separ}
\end{figure}

All solutions in Group 2 are one \textbf{subfunction modification} away from the solution in Group 1: that is, the value of exactly one subfunction [from the three separable subfunctions] is different. Every solution in Group 2 is one modification away from \textbf{some} of the solutions in Group 3. Finally, all solutions in Group 3 are one modification away from the optimal solution in Group 4. This example shows that we can start from any solution, greedily optimise it, and (if we know the problem structure) it is enough to optimise only one subfunction at a time to get to the optimal solution. Thus, the LON we wish to get is presented in Figure \ref{fig:separ}.

Consider an overlapping problem $f_{ovr}(x_0,...,x_9) = f_1(x_0,...,x_3)+f_2(x_3,...,x_6) + f_3(x_6,...,x_9)$, where each $f_s(\vec{x}_{I_s})=bimTrap_4(u(\vec{x}_{I_s}))$. Each subfunction in $f_{ovr}$ shares one or two variables with its neighbours. Thus, the modification of shared variables influences more than one subfunction. Optimal solutions to $bimTrap_4$ are $0000$ and $1111$. Locally optimal solutions contain two zeros and two ones, e.g., $0110$ and $1001$.
\begin{figure}[h]
    \centering
    \includegraphics[width=0.99 \linewidth]{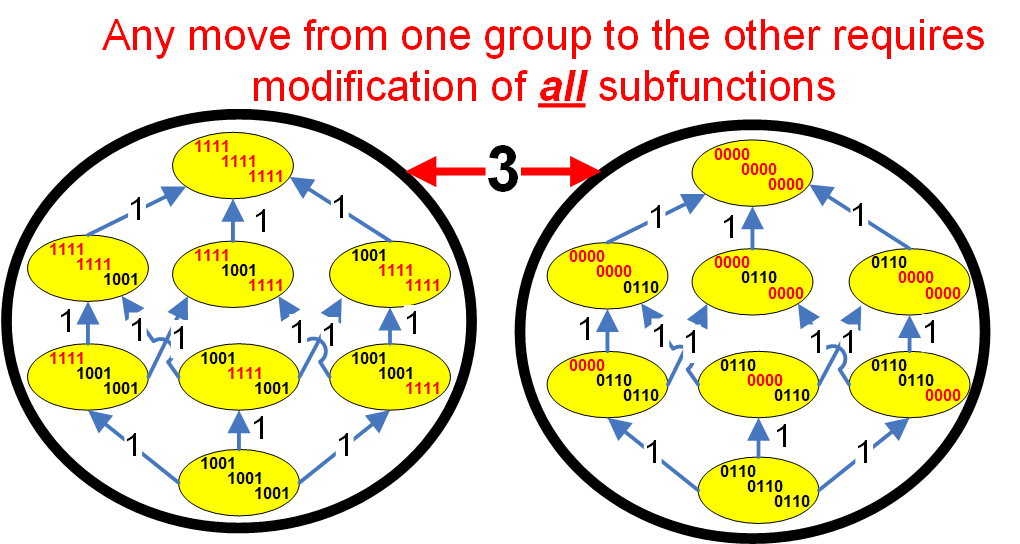}
    \caption{Decomposition-aware LON for overlapping problem}
    \label{fig:overlap}
\end{figure}

In Figure \ref{fig:overlap}, we present a LON, which concerns two locally optimal solutions as its roots. The left LON part starts from the solution in which the arguments of each subfunction are $1001$, while in the right part of a LON, the arguments of each subfunction are $0110$. Same as in Figure \ref{fig:separ}, we consider the improvement of a single subfunction. The optimal solutions (built only from '1's or only from '0's) are located on the top of the two LON parts. In a jump from any solution on the left part of the LON to any solution on the right part of the LON, all subfunctions must be modified. Such a situation would not occur if the subfunction did not overlap. Thus, the overlaps have caused the creation of two separate attraction basins of the two global optima. Moving between these two attraction basins is hard because it requires the modification of all subfunctions. In the latter part of this work, we show that we can observe such separate attraction basins in the proposed decomposition-aware LONs obtained experimentally.

\section{Methods}

\subsection{LON Construction}

\paragraph{Traditional (black-box) ILS} Hereafter referred to as algorithm \A{}. In accordance with LON literature \cite{ochoa2017understanding,thomson2023randomness,mostert2019insights}, we base the traditional construction algorithm on sampling using iterated local search (ILS). Several independent runs of ILS are executed; these follow cycles of perturbation followed by local search. In our case, the mutation operator is a single bitflip. Perturbation is three bitflips, and the local search is first-improvement --- accepting strictly improving solutions. For the local optimum acceptance criterion, improving or equal local optima are accepted. 

\begin{figure*}[ht!]
\centering
\begin{subfigure}[b]{0.33\textwidth}   
\centering 
\includegraphics[trim = 70 60 0 70, clip,width=\textwidth]{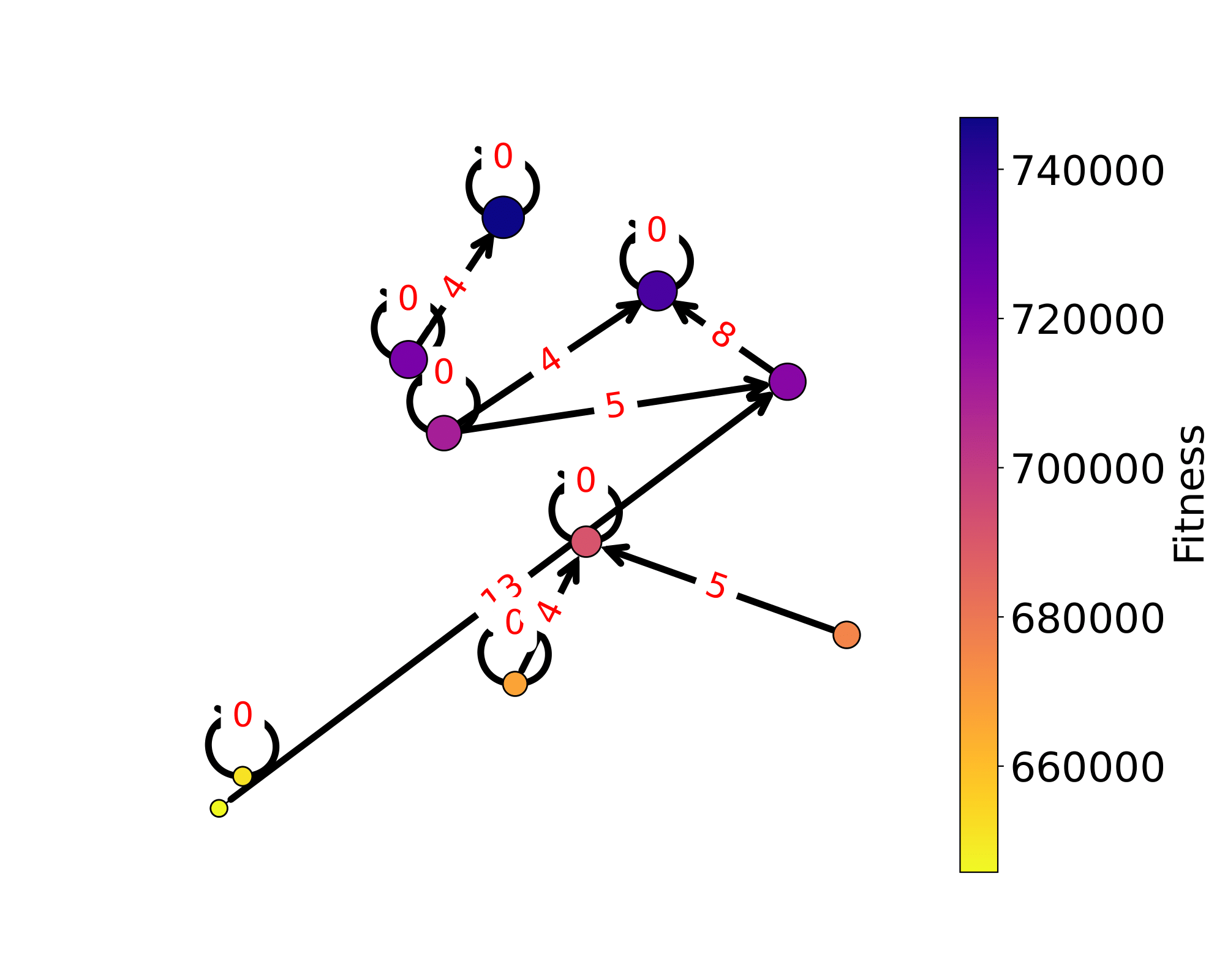}
\caption{Algorithm \A{}}\label{fig:lon-a}
\end{subfigure}
\begin{subfigure}[b]{0.33\textwidth}   
\centering 
\includegraphics[trim = 70 60 0 70, clip, width=\textwidth]{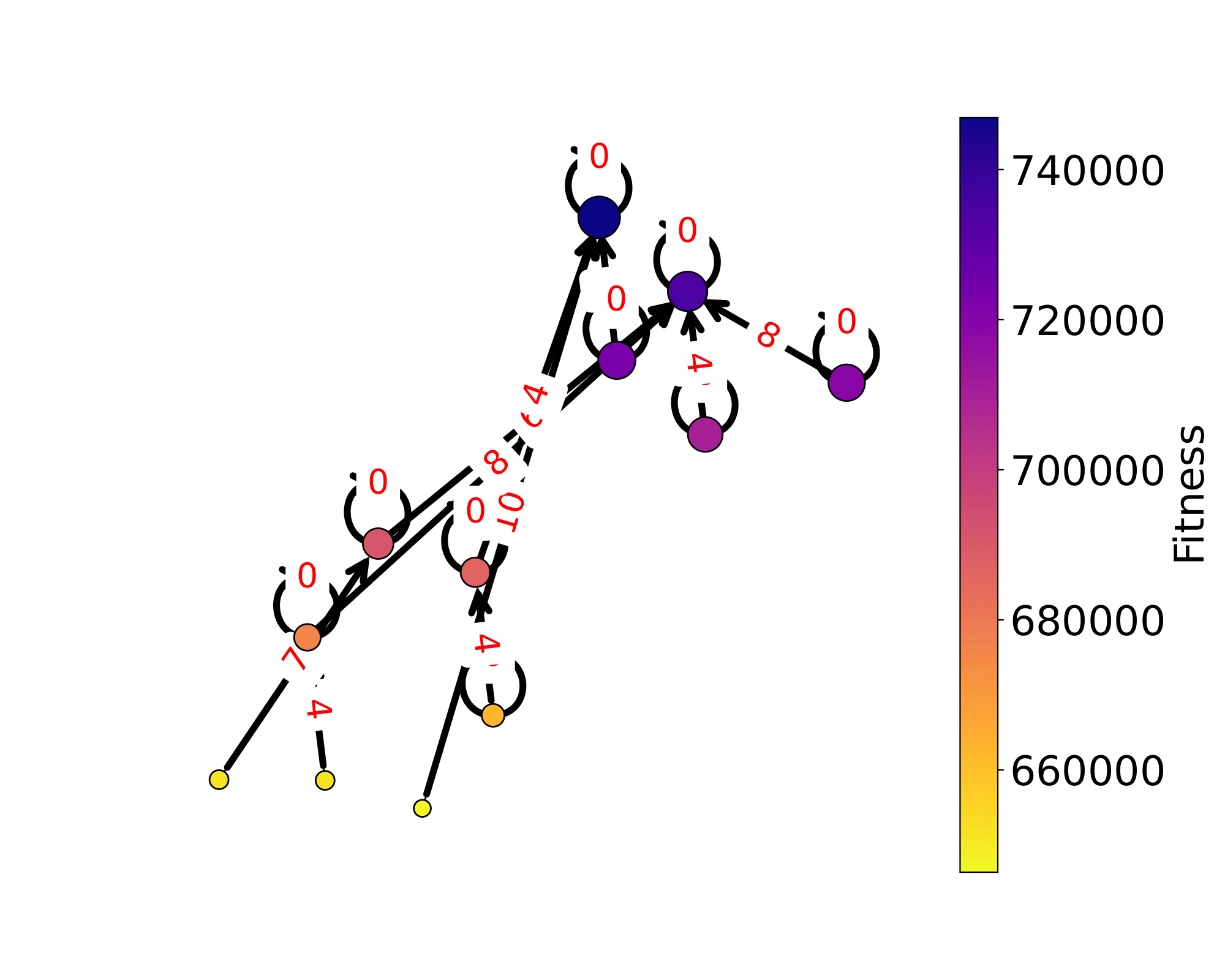}
%includegraphics[width=0.7 \linewidth]{images/emptyImage.png}
\caption{Algorithm \B{}}\label{fig:lon-b}
\end{subfigure}
\begin{subfigure}[b]{0.33\textwidth}   
\centering 
\includegraphics[trim = 70 60 0 70, clip,width=\textwidth]{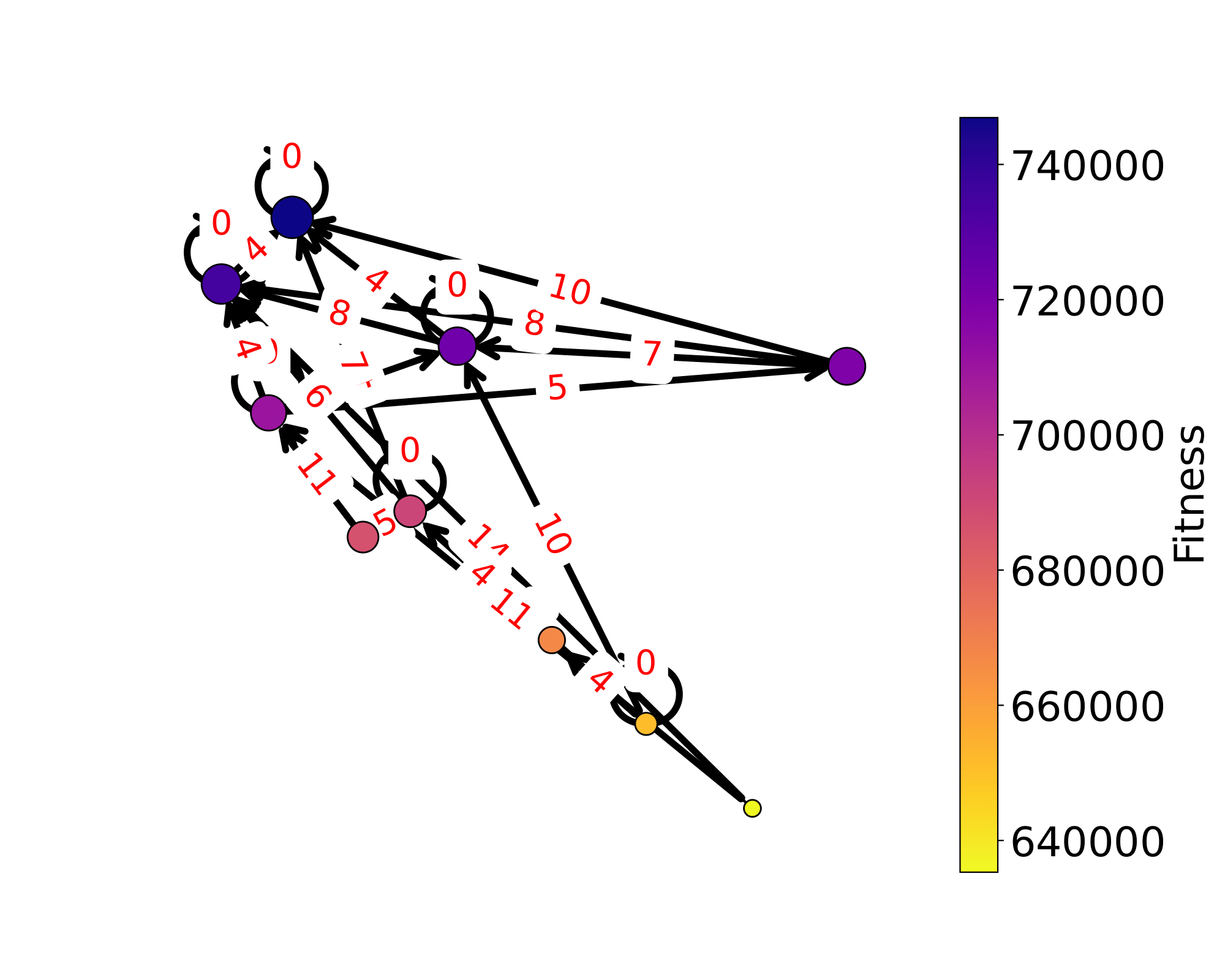}
\caption{Algorithm \C{}}\label{fig:lon-c}
\end{subfigure}
\caption{Local optima networks for an NK landscape of size 15 with $k$ = 2. Node size, colour, and position on the $y$-axis is relative to fitness (maximisation). Position on the $x$-axis is obtained through multi-dimensional scaling of the binary solutions using Hamming distance}\label{fig:lon-viz}
\end{figure*}

\sloppypar{\paragraph{Deterministic recombination ILS} Hereafter referred to as algorithm \B{}. Instead of perturbation chaining together the local searches, partition crossover [as described in Section \ref{sec:rw:varDeps}] is used. This algorithm is from a previous work \cite{chicano2017optimizing} except we use ordinary first-improvement local search instead the specialised speedup hill-climber employed in that paper [they were optimising very high-dimensional problems]. The approach has been called deterministic recombination iterated local search (DRILS) in previous literature.}

\paragraph{VIGp with FIHCwLL} Hereafter referred to as algorithm \C{}. Based on the approach in a previous work \cite{tinos2022iterated}, this algorithm uses VIGp as described in Section \ref{sec:rw:varDeps} as the perturbation operator. We use first improvement hill climber with linkage learning (FIHCwLL) \cite{FIHCwLL} to locally optimise solutions and learn dependencies simultaneously. FIHCwLL optimises a solution by flipping its genes. If a gene flip improves fitness, then it is preserved, or it is rejected. In each iteration, FIHCwLL considers all genes in a random order. Iterations are executed until at least one gene is modified during the preceding iteration. Fitness evaluations that are used for local optimisation are also used to perform variable dependency checks. Since we consider \textit{k}-bounded problems that are additively decomposable, we use the non-linearity dependency check (Formula \ref{eq:nonLinear}) that requires performing four fitness evaluations, i.e.,  $f(\vec{x})$, $f(\vec{x}^g)$, $f(\vec{x}^h)$,  and $f(\vec{x}^{g,h})$. In FIHCwLL, three of them are the side-effect of the local search, and only one must be computed for linkage discovery purposes, which significantly reduces the cost of decomposition. FIHCwLL was inspired by \cite{ilsDLED}.

\subsection{LON metrics}

\paragraph{Subfunction structure.} We propose metrics relating to subfunction structure in LONs. These can be described as follows:

\begin{itemize}
    \item number of subfunction changes associated with a LON directed edge between two local optima: that is, the number of subfunctions which changed when comparing the destination node to the source node
    \begin{itemize}
        \item a LON typically contains several edges, so we consider the mean, median, and standard deviation of this metric across the network
    \end{itemize}
    \item the number of positive subfunction changes in LON edges
    \begin{itemize}
        \item for a decision problem such as MAX3SAT, this is the number of subfunctions which changed from unsatisfied in the source node to satisfied in the destination node
        \item for an optimisation problem such as NK landscapes, this is the number of subfunctions which increased in value (assuming maximisation) from the source to the destination node
        \item the mean, median, and standard deviation of this metric across a LON are considered
    \item the number of negative subfunction changes in LON edges
    \begin{itemize}
        \item this is the same as just described, except for negative changes to subfunction values
    \end{itemize}
    \end{itemize}
\end{itemize}

\paragraph{Previous metrics.} In addition to the subfunction metrics, we also compute 14 LON metrics from previous literature: \cite{ochoa2017understanding,herrmann2018pagerank,thomson2023randomness}. None of these aim at capturing subfunction structure, but we would nevertheless like to compare them and ascertain whether they capture different information. Two of the metrics relate to the number of local and global optima; two describe neutrality at the local optima level; nine are related to the notion of landscape \emph{funnels} [basins of attraction at the local optima level]; and one considers the pagerank centrality of the global optimum. 

\section{Experimental Setup}
\subsection{Problem instance generation} 

We consider a set of well-known benchmarks that include NK-landscapes, MAX3SAT, and deceptive functions. All instances are of toy size: 15 bits for NK-landscapes and MAX3SAT, and between 15 and 18 bits for the deceptive functions due to varying subfunction structure setups which necessitate different problem sizes. Thus, it is easy to analyse the obtained LONs in terms of their expected and obtained features. For NK-landscape and MAX3SAT generation, we have adopted the software from \cite{P3Original} and generated 30 instances each. For the deceptive functions, there is no element of randomness in their construction. We consider 11 different problems of this type, with varying numbers of subfunctions and degrees of subfunction overlap. All instances used in this work are available in the supplemental material [see footnote \ref{fn:supp}]. 

\subsection{LON Construction}

For all three LON construction algorithms, there are 30 independent runs per problem instance. ILS runs terminate when there has not been a [strict] improvement to local optimum quality in 30 full cycles. These numbers are sufficient given the search space sizes under study, which comprise a maximum of \(2^{18}\) = 262,144 solutions. LON construction and analysis are coded from scratch in Python. 

\subsection{Visualisation}

The LON visualisations are implemented using the Python libraries \textsc{NetworkX} \cite{hagberg2020networkx} and \textsc{Matplotlib} \cite{ari2014matplotlib} For all network visualisations shown, the position of a node on the $x$-axis is the result of applying multi-dimensional scaling (MDS) to the local optima sample, using Hamming distance between the binary vectors as the measurement of distance. \textsc{scikit-learn} \cite{pedregosa2011scikit} is used for MDS and \textsc{SciPy} \cite{virtanen2020scipy} for the distance computations.

\begin{figure*}[ht!]
\centering
\begin{subfigure}[b]{0.7\textwidth}  
    \includegraphics[width=\textwidth]{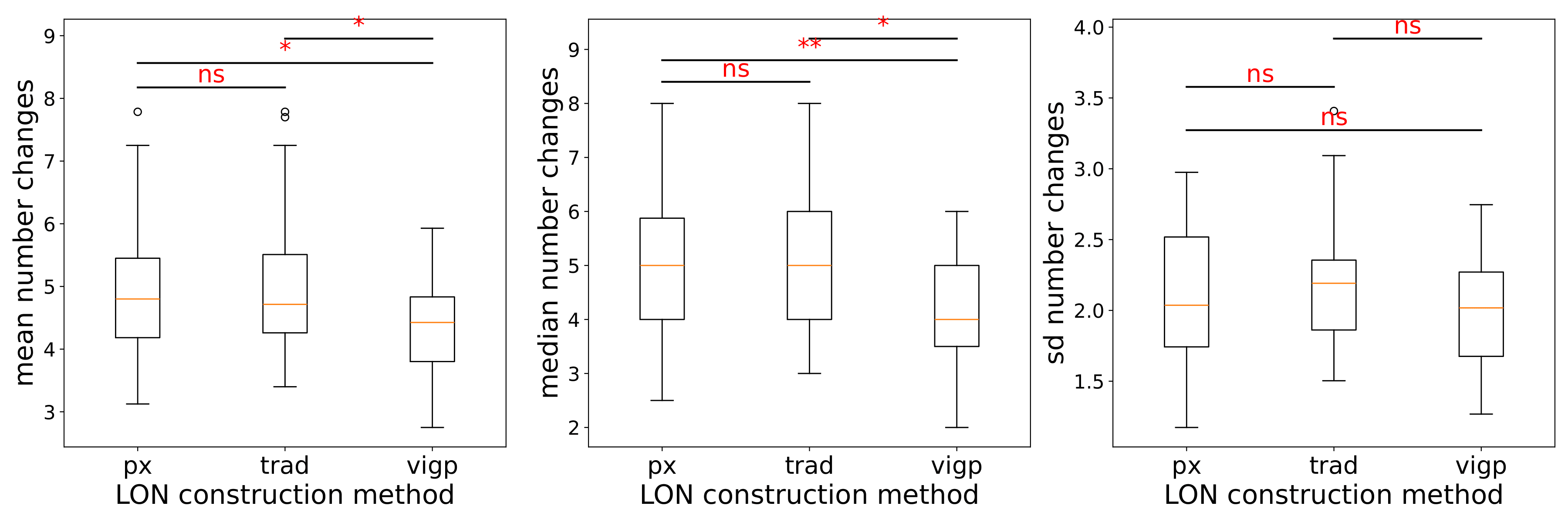}
\end{subfigure}
\caption{Distribution of LON edge subfunction metrics for 30 MAX3SAT instances [clauses-to-variable ratio 4.27]. Metrics consider the number of subfunctions which change in value between the start and end of LON edges [for improving edges only]. An indication of significant difference between pairs according to a Mann-Whitney test is annotated on the plots; ns means not significant; * means $p<0.05$, ** means $p<0.01$, *** means $p<0.001$}\label{fig:subfunc-metrics-1}
\end{figure*}

\begin{figure*}[ht!]
\centering
\begin{subfigure}[b]{\textwidth}  
    \includegraphics[trim = 0 300 0 0, clip,width=0.98\textwidth]{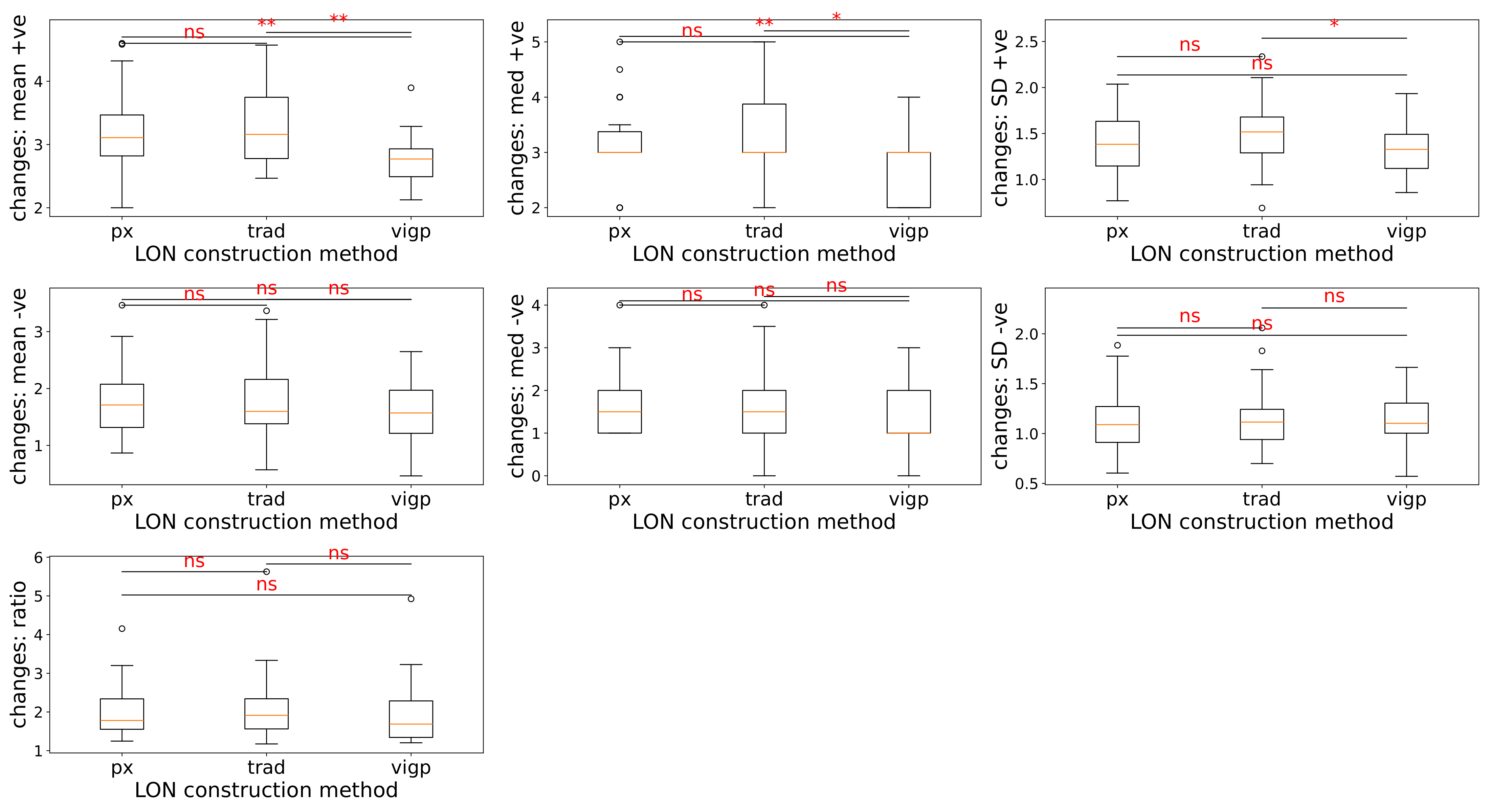}
\end{subfigure}
\caption{Distribution of LON edge subfunction metrics for 30 MAX3SAT instances [clauses-to-variable ratio 4.27]. The metrics consider the number of positive (+ve) and negative (-ve) changes between the start and end of LON edges [for improving edges only]. Indication of significant difference between pairs according to a Mann-Whitney test is annotated on the plots: ns means not significant; * means $p<0.05$, ** means $p<0.01$, *** means $p<0.001$}\label{fig:subfunc-metrics-2}
\end{figure*}

\begin{figure*}[ht!]
\centering
\begin{subfigure}[b]{0.48\textwidth}   
\centering 
\includegraphics[trim = 0 0 0 18, clip,width=\textwidth]{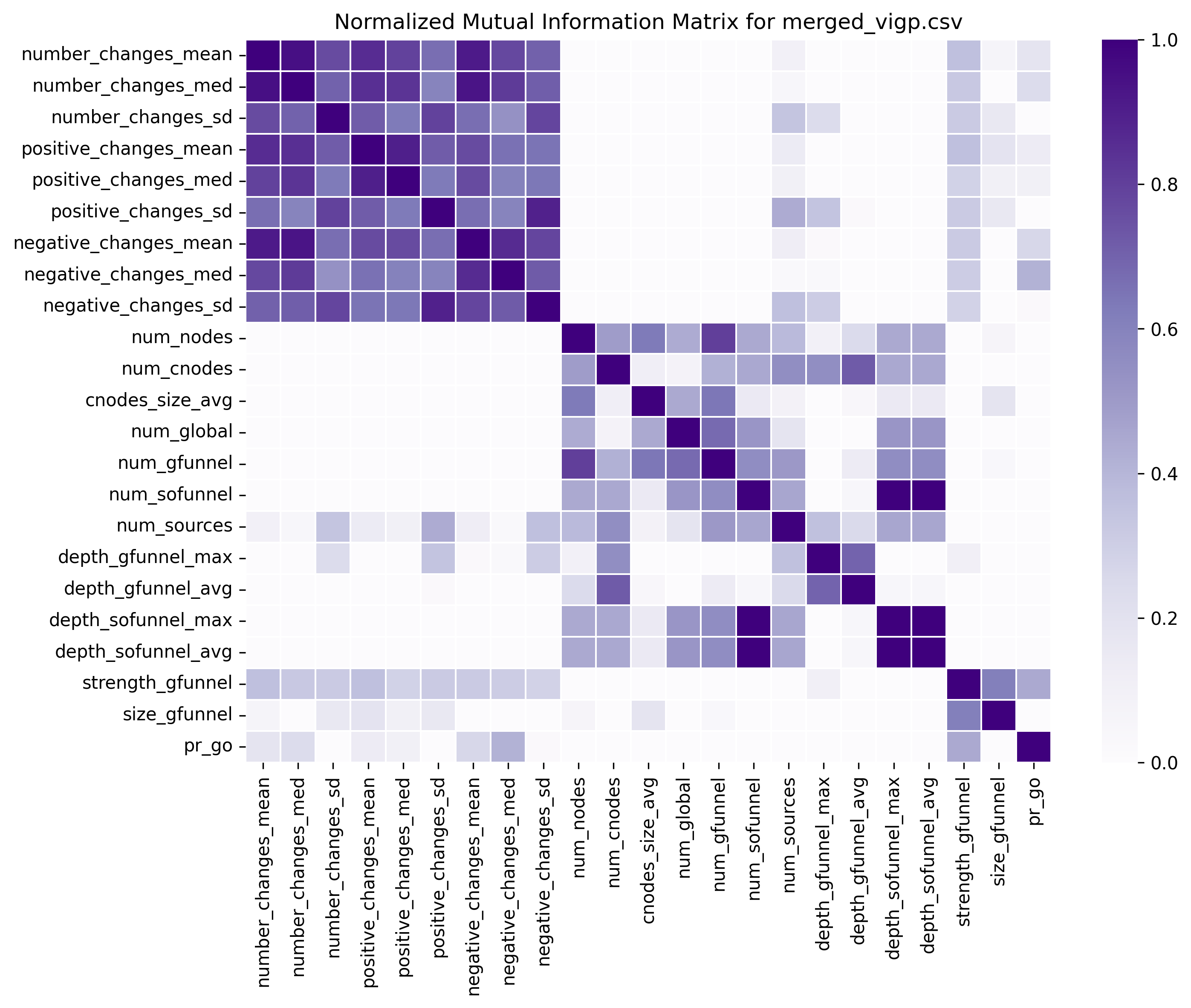}
\caption{Kendall-tau correlation}\label{fig:kt}
\end{subfigure}
\begin{subfigure}[b]{0.48\textwidth}   
\centering 
\includegraphics[trim = 0 0 0 18, clip, width=\textwidth]{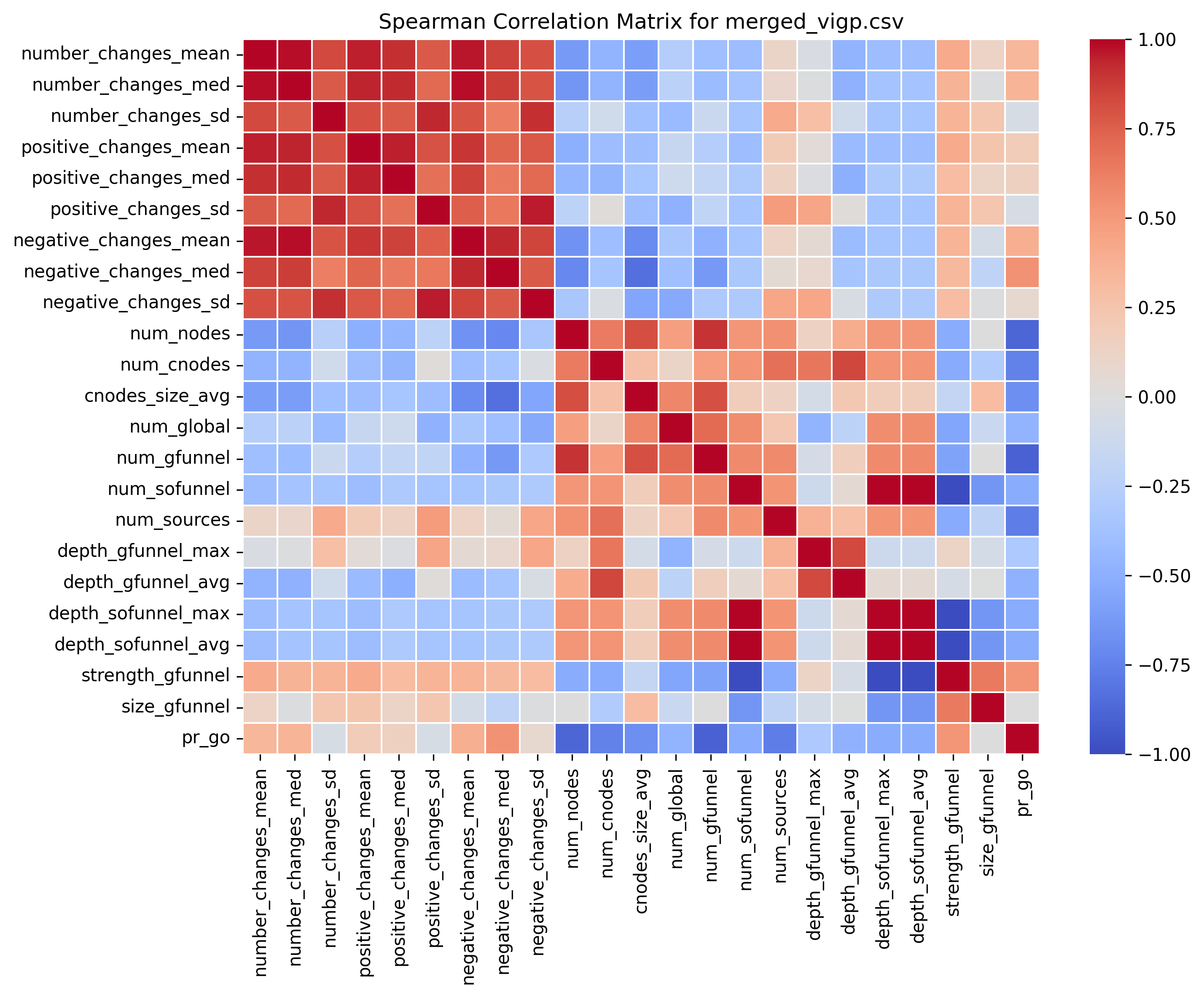}
\caption{Spearman correlation}\label{fig:spearman}
\end{subfigure}
\caption{Similarity between LON metrics (algorithm \C{}) for deceptive functions. The proposed LON subfunction metric names contain the substring "changes"; all other metrics are from previous LON literature}\label{fig:correlations}
\end{figure*}

\begin{figure*}[ht!]
\centering
\begin{subfigure}[b]{0.33\textwidth}   
\centering 
\includegraphics[width=\textwidth]{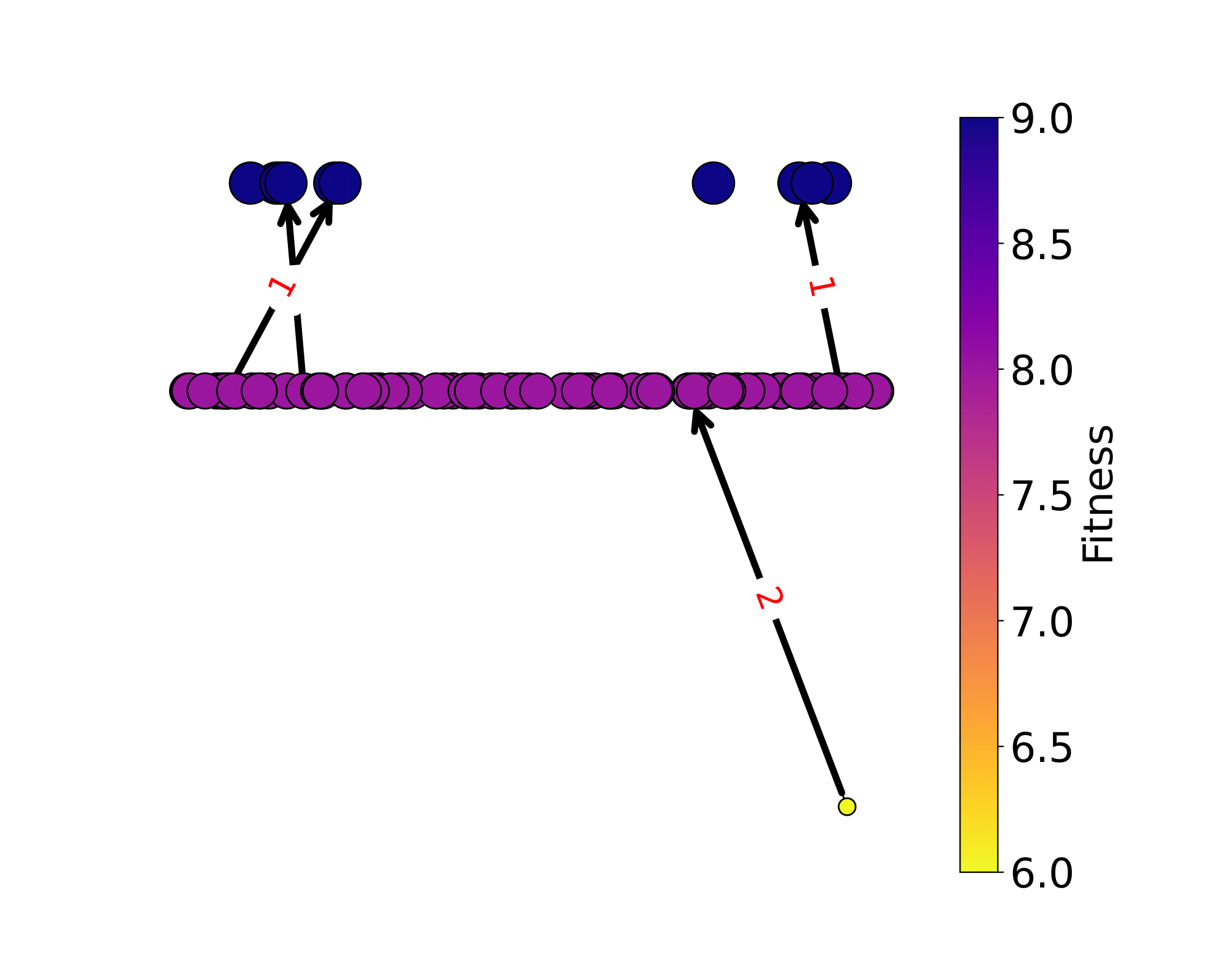}
\caption{Subfunction overlap of 2; Algorithm \A{}}\label{fig:trad-2}
\end{subfigure}
\begin{subfigure}[b]{0.33\textwidth}   
\centering 
\includegraphics[width=\textwidth]{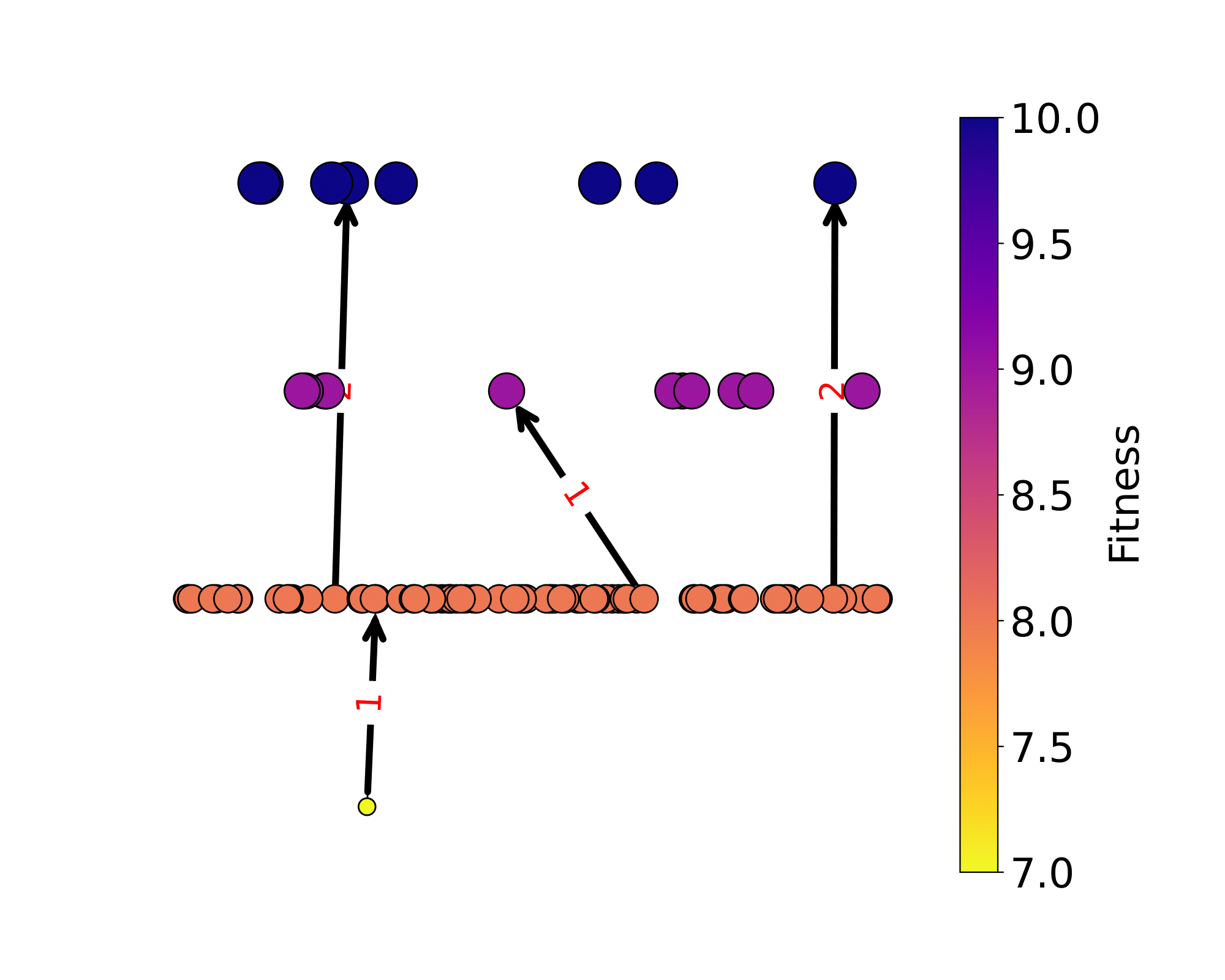}
\caption{Subfunction overlap of 2; Algorithm \B{}}\label{fig:px-2}
\end{subfigure}
\begin{subfigure}[b]{0.33\textwidth}   
\centering 
\includegraphics[angle=270,width=\textwidth]{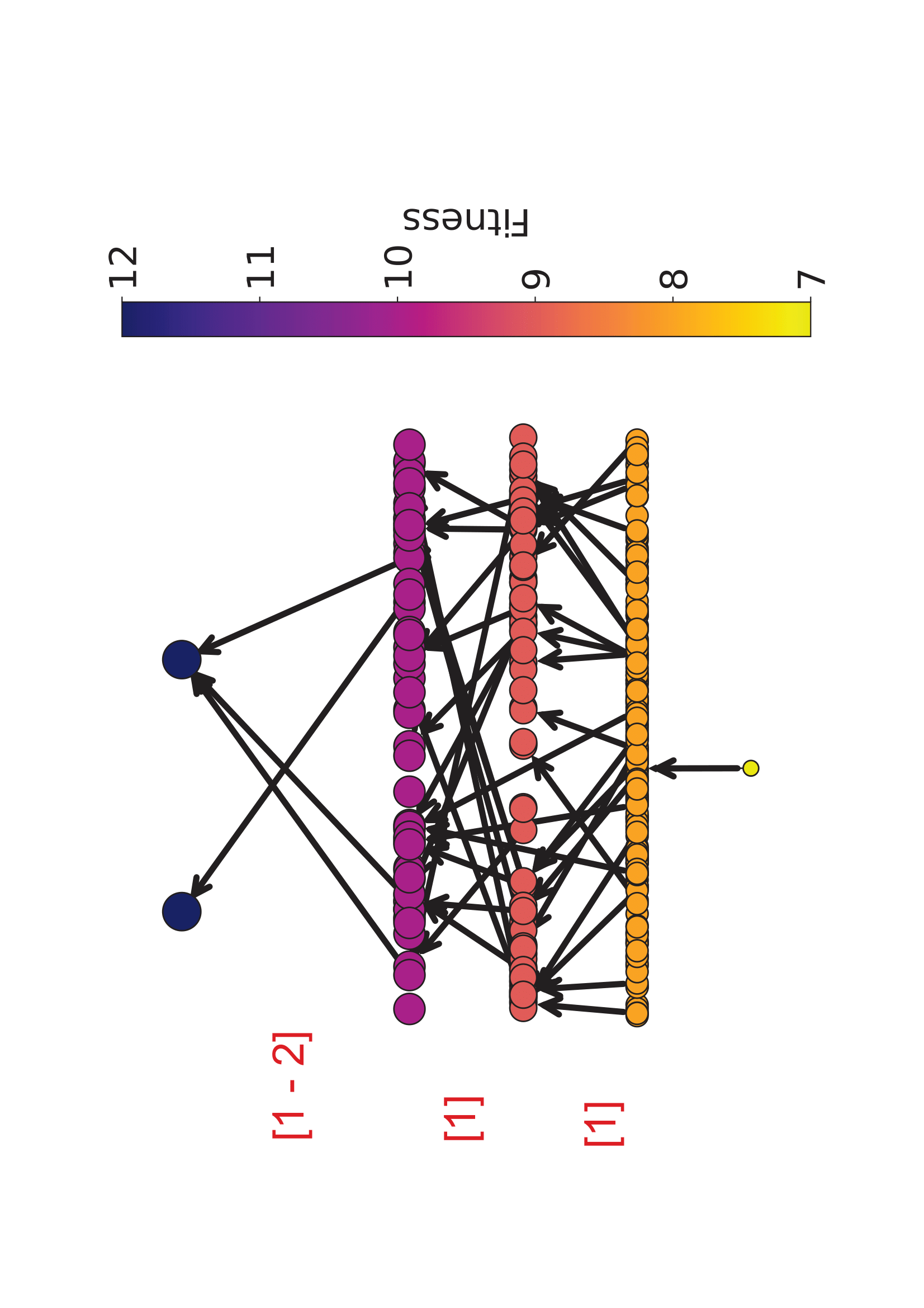}
\caption{Subfunction overlap of 2; Algorithm \C{}}\label{fig:vigp-2}
\end{subfigure}
\begin{subfigure}[b]{0.33\textwidth}   
\centering 
\includegraphics[angle=270,width=\textwidth]{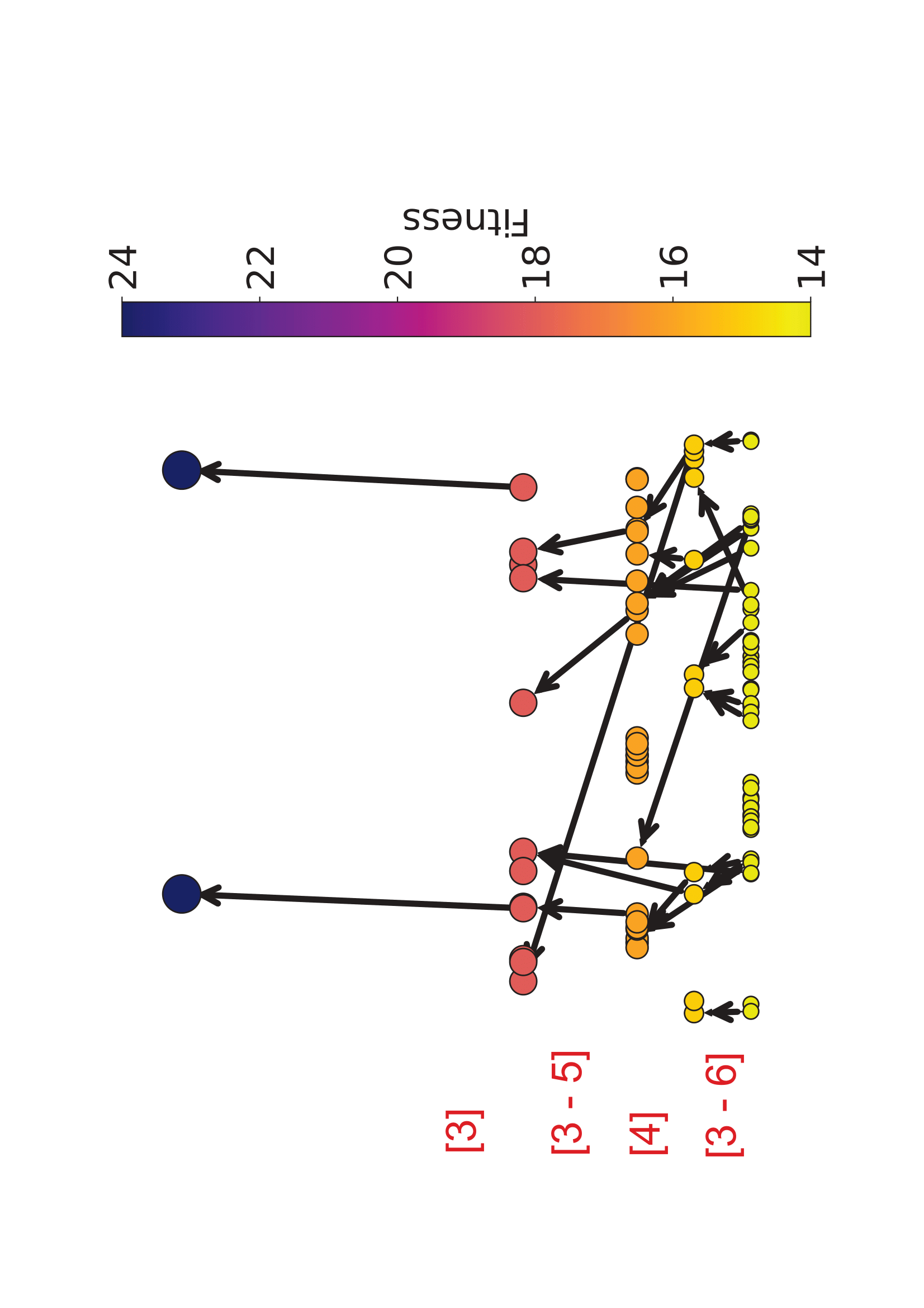}
\caption{Subfunction overlap of 4; Algorithm \A{}}\label{fig:trad-4}
\end{subfigure}
\begin{subfigure}[b]{0.33\textwidth}   
\centering 
\includegraphics[trim = 20 20 20 20, clip, angle=270,width=\textwidth]{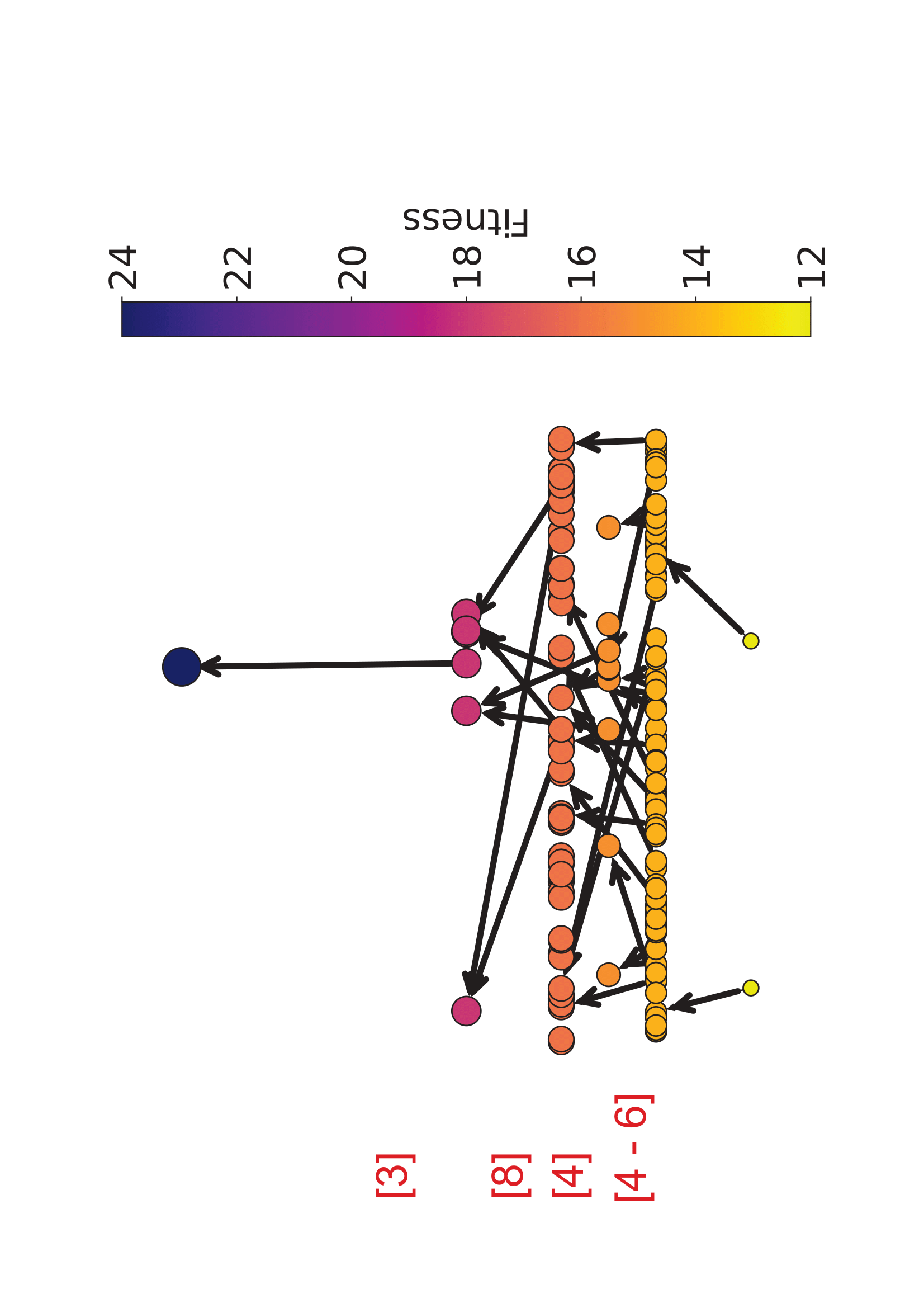}
\caption{Subfunction overlap of 4; Algorithm \B{}}\label{fig:px-4}
\end{subfigure}
\begin{subfigure}[b]{0.33\textwidth}   
\centering 
\includegraphics[angle=270,width=\textwidth]{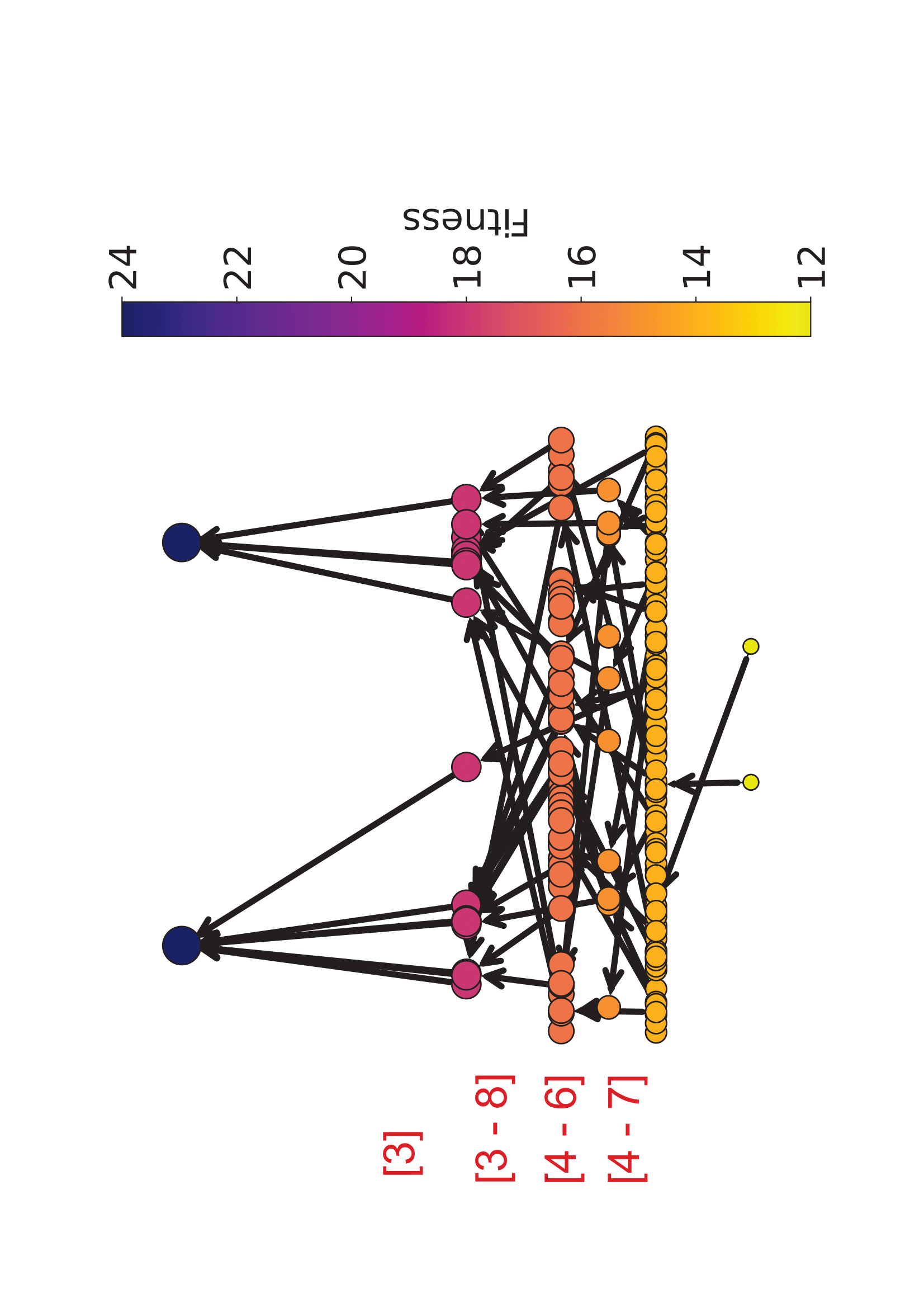}
\caption{Subfunction overlap of 4; Algorithm \C{}}\label{fig:vigp-4}
\end{subfigure}
\caption{Local optima networks for two deceptive problems with different degrees of subfunction overlap. Node size, colour, and position on the $y$-axis is relative to fitness (maximisation). Red text on an individual edge is the number of subfunction changes between the source and destination node. Red text in brackets indicates the range of subfunction changes needed to jump between the two fitness levels beside it. Position on the $x$-axis is obtained through multi-dimensional scaling of the binary solutions using Hamming distance}\label{fig:lon-overlap}
\end{figure*}

\section{Results}
\subsection{Network visualisation}
Figure \ref{fig:lon-viz} shows local optima networks for an NK landscape instance constructed using the three different algorithms. Note that visualisations for all other algorithms and problems under study are available in the supplemental material \footnote{\bulurl{https://zenodo.org/records/14767046}\label{fn:supp}}. In the Figure, each node is a local optimum and each edge is a directed transition between them. Node size, colour, and position on the $y$-axis are relative to fitness (maximisation). Position on the $x$-axis is the result of multi-dimensional scaling on the solution sample. The red numbers annotated onto edges capture the number of subfunctions which changed in the transition from the source local optimum to the destination local optimum. 

Comparing the three sub-plots, we can immediately notice that there is a difference between Algorithm \C{} (Figure \ref{fig:lon-c}) and the two others. While Algorithms \A{} and \B{} lend to rather sparse and simple networks, Algorithm \C{} leads to a denser view with more information. Notice that although all three reach the same fitness level, Algorithms \A{} identifies only a single edge leading towards the solution with this fitness. On the other hand, Algorithms \B{} and \C{} reveal three and four edges (respectively) directed towards the highest fitness level. Looking at the edge labels, it seems that these transitions require quite dramatic subfunction alterations in the solution: as many as 8-10 changes. It is interesting that these two algorithm were able to identify the dramatic changes which are needed to ascend. 

\subsection{LON subfunction metrics}

Next we consider the distribution of metrics related to subfunction changes associated with edges of a LON. For every directed edge we assess which subfunctions changed in value; for those, we consider the number and \emph{direction} of those changes: did the subfunction value increase (positive) or decrease (negative)? Figure \ref{fig:subfunc-metrics-1} shows, for 30 MAX3SAT instances with a clause-to-variable ratio of 4.27: the mean, median, and standard deviation number of subfunction changes encoded in LON edges. Results are shown for the three construction algorithms, and indications of statistical difference betwween pairs of distributions are annotated, as described in the caption. 

Looking across the plots in Figure \ref{fig:subfunc-metrics-1}, we can see that with respect to mean and median number of changes Algorithm \C{} appears to have significantly different distributions when compared to Algorithms \A{} and \B{}. From visual inspection of the boxes we can see that the Algorithm \C{} distributions are lower than those of the other two algorithms (that is, there are less subfunction changes in the directed edges between two local optima). Algorithms \A{} and \B{} have no significant difference in these two metrics. With respect to the standard deviation (the right-most plot), none of the algorithms differ significantly from one another. 

Figure \ref{fig:subfunc-metrics-2} has the same layout as Figure \ref{fig:subfunc-metrics-1} but presents different metrics. Instead of the number of subfunction changes, these measurements consider the \emph{direction}: that is, how many subfunctions a.k.a clauses changed from not satisfied to satisfied ---- a positive change --- and how many changed from satisfied to not satisfied --- a negative change ---- between the source and destination of a LON edge. We notice that for the positive changes, there is significant difference between Algorithm \C{} and the two others with respect to the mean and median. Algorithm \C{} makes less positive changes; however, we can recall from Figure \ref{fig:subfunc-metrics-1} that this algorithm makes less changes overall. Such an observation indicates that it makes smaller steps in terms of the number of the modified subfunctions. Such a feature may be considered important because VIGbp can be considered as making more precise steps. Therefore, VIGbp-generated LONs are more dense, and VIGbp visits more local optima than PX. Note that considering smaller variation masks that allow the improvement makes finding this improvement easier. Therefore, some studies focus on limiting the variation of mask sizes \cite{GoldMonotonicity,superMasksGray}. There is no significant difference between the algorithms with respect to negative changes.  

Plots for the other problems are available in the supplemental material [see footnote \ref{fn:supp}]. In the case of NK landscapes, Algorithm \B{} makes less subfunction changes than the other two algorithms, and this difference has statistical significance. Algorithms \A{} and \C{} make more positive changes than Algorithm \B{} --- although this may be an artefact of the fact that they make more changes in general. In terms of negative subfunction changes, Algorithm \C{} makes more of them when compared with the other to algorithms [with indication of statistical significance]. The deceptive problem group, on the other hand, is not associated with significant differences between the three algorithms. This may be due to the low number of problems in the group [10, because one of the problems has LONs with no edges and therefore no edge subfunction metrics].

\subsection{LON metric comparison}
We would like to consider to what extent the LON subfunction metrics proposed here are similar to different, previously-proposed metrics for LONs. To this end, Figure \ref{fig:correlations} presents, for the deceptive problem set, the Kendall-Tau correlation and also the Spearman correlation for pairs of LON measurements. LON metrics are named on each axis [the proposed subfunction metric names contain {\lq{changes}\rq} and the colour in a square captures the similarity metric, as indicated in the colourbar. 

Looking across Figure \ref{fig:kt}, we can notice from looking along the rows associated with the subfunction metrics that while they have moderate-to-strong correlations with each other, they have only weak or very weak correlation to other LON metrics from the literature. The Spearman correlations in Figure \ref{fig:spearman} show that the subfunction metrics have moderate-to-strong correlations among themselves, and weak-to-moderate correlations with other LON metrics from the literature. In most cases this is a [weak] negative correlation, but there are a few metrics from previous works which show [weak] positive correlations with the subfunction measurements. Correlation matrices for the other problems and algorithms can be found in the supplemental material [footnote \ref{fn:supp}]; these showed similar trends to those just described. 

\subsection{Case study: subfunction overlap}
To demonstrate the insights that can be gained using our approach, we consider a case study where we would like to analyse the difference in landscape structure when problems have a different degree of subfunction overlap. To this end, we choose two bimodal deceptive problems for comparison: one has subfunctions with a variable overlap of two, and the other has an overlap of four. LONs for both of them, constructed using the three algorithms, are presented in Figure \ref{fig:lon-overlap}. In these plots, red text on an individual edge represents the number of subfunction changes between the source and destination node. Red text in brackets indicates the range of subfunction changes needed to jump between the two fitness levels beside it. 

Looking at the first row of plots, which relates to the problem with lower subfunction overlap, we can observe that only Algorithm \C{} found the global optimum fitness (which is 12). In fact, it identifies both of the global optima. The Algorithm \A{} and Algorithm \B{} LONs are very sparse --- both of them have only four edges. On the other hand, Algorithm \C{} is a densely-connected network, with plenty of opportunities for search to transition between fitness levels. From the red text annotations, we can see that moving up one fitness level is associated with a low number of subfunction changes: typically one and maximum two. Thus, thanks to using smaller steps, it is easier for \C{} to traverse the network of local optima. This observation is coherent with the aforementioned research direction to eliminate dependencies irrelevant to optimisation and obtain shorter variation masks \cite{GoldMonotonicity,superMasksGray}. Indeed, Algorithm \C{} can find a path to a global optimum from almost any initial state (e.g., Fig \ref{fig:vigp-4}). Oppositely, for \A{}- and \B{}-based LONS, the number of paths from a randomly chosen state to global optimum is low (e.g., Fig \ref{fig:trad-4}) and \ref{fig:trad-4}.

Looking now at the lower row of plots, which are for the instance with a higher degree of subfunction overlap. When compared to the problem with lower subfunction overlap, both Algorithm \A{} and Algorithm \B{} have an increased number of connections in Figure \ref{fig:trad-4} and \ref{fig:px-4}. Comparing the top row of plots with the bottom row, we can notice that the problem with higher subfunction overlap has LONs with more dramatic subfunction changes between fitness levels. On the higher-overlap problem (Figures \ref{fig:trad-4}-\ref{fig:vigp-4}), all of the three algorithms reach at least one global optimum (fitness of 24). Algorithm \B{} reaches only one global optimum and there is a single connection towards it. Algorithms \A{} and \C{} reach both global optima, but we notice that the Algorithm \C{} LON contains several connections towards them, while the Algorithm \A{} LON has only two. Another interesting observation is that Algorithms \B{} and \C{} find connections which require eight subfunctions to change between the second and third-highest fitness levels. Algorithm \A{} did not find these options and this may be a reason for the sparse connectivity of its LON in relation to the global optima.

\section{Conclusion}
We have considered whether including problem structure information into local optima network (LON) construction and analysis can bring additional insights into optimisation dynamics. Three well-known $k$-bounded problem domains were included in the analysis: MAX3SAT, NK landscapes, and deceptive bimodal problems. We constructed LONs in different ways: the standard {\lq{black box}\rq} approach of iterated local search (ILS) runs; an ILS algorithm using partition crossover which incorporated subfunction structure information a-priori; and an ILS algorithm which learns and uses subfunction structure during the search. The LONs were visualised in a new manner which conveys the number of subfunction changes between local optima. New metrics were proposed: these relate to the number and direction of subfunction changes between optima. These were compared across the three LON algorithm approaches and also compared against existing LON metrics from the literature. 

The results showed considering subfunction structure in the analysis of LONs can increase the amount of insight gained into optimisation dynamics: for example, identifying how many subfunctions must change in order to ascend fitness levels. The proposed metrics were shown to contain different information to existing LON metrics from the literature. Lastly, a case study was presented. We used the new approach to compare landscape structure and algorithm trajectories between problems with differing degrees of subfunction overlap. Future work will consider the scalability of our approach, as well as comparing the LON subfunction structure analysis with an analysis done using only partial learned problem information. Code, data, and additional plots associated with this work are available in a dedicated Zenodo repository [see footnote \ref{fn:supp}].

\paragraph{Acknowledgements.} The work of Michal W. Przewozniczek was supported by the Polish National Science Centre (NCN) under Grant 2022/45/B/ST6/04150.

\balance

	\bibliographystyle{ACM-Reference-Format}
	\bibliography{rootDoc} 


\begin{thebibliography}{44}

%%% the Web address.

\ifx \showCODEN    \undefined \def \showCODEN     #1{\unskip}     \fi
\ifx \showISBNx    \undefined \def \showISBNx     #1{\unskip}     \fi
\ifx \showISBNxiii \undefined \def \showISBNxiii  #1{\unskip}     \fi
\ifx \showISSN     \undefined \def \showISSN      #1{\unskip}     \fi
\ifx \showLCCN     \undefined \def \showLCCN      #1{\unskip}     \fi
\ifx \shownote     \undefined \def \shownote      #1{#1}          \fi
\ifx \showarticletitle \undefined \def \showarticletitle #1{#1}   \fi
\ifx \showURL      \undefined \def \showURL       {\relax}        \fi
% invisible to TeX
\providecommand\bibfield[2]{#2}
\providecommand\bibinfo[2]{#2}
\providecommand\natexlab[1]{#1}
\providecommand\showeprint[2][]{arXiv:#2}

\bibitem[Ari and Ustazhanov(2014)]%
        {ari2014matplotlib}
\bibfield{author}{\bibinfo{person}{Niyazi Ari} {and} \bibinfo{person}{Makhamadsulton Ustazhanov}.} \bibinfo{year}{2014}\natexlab{}.
\newblock \showarticletitle{Matplotlib in python}. In \bibinfo{booktitle}{\emph{2014 11th International Conference on Electronics, Computer and Computation (ICECCO)}}. IEEE, \bibinfo{pages}{1--6}.
\newblock


\bibitem[Bouter et~al\mbox{.}(2018)]%
        {partialAnton}
\bibfield{author}{\bibinfo{person}{Anton Bouter}, \bibinfo{person}{Tanja Alderliesten}, \bibinfo{person}{Arjan Bel}, \bibinfo{person}{Cees Witteveen}, {and} \bibinfo{person}{Peter A.~N. Bosman}.} \bibinfo{year}{2018}\natexlab{}.
\newblock \showarticletitle{Large-scale parallelization of partial evaluations in evolutionary algorithms for real-world problems}. In \bibinfo{booktitle}{\emph{Proceedings of the Genetic and Evolutionary Computation Conference}} (Kyoto, Japan) \emph{(\bibinfo{series}{GECCO '18})}. \bibinfo{publisher}{Association for Computing Machinery}, \bibinfo{address}{New York, NY, USA}, \bibinfo{pages}{1199–1206}.
\newblock
\showISBNx{9781450356183}
\href{https://doi.org/10.1145/3205455.3205610}{doi:\nolinkurl{10.1145/3205455.3205610}}


\bibitem[Canonne et~al\mbox{.}(2023)]%
        {canonne2023combine}
\bibfield{author}{\bibinfo{person}{Lorenzo Canonne}, \bibinfo{person}{Bilel Derbel}, \bibinfo{person}{Francisco Chicano}, {and} \bibinfo{person}{Gabriela Ochoa}.} \bibinfo{year}{2023}\natexlab{}.
\newblock \showarticletitle{To Combine or not to Combine Graybox Crossover and Local Search?}. In \bibinfo{booktitle}{\emph{Proceedings of the Genetic and Evolutionary Computation Conference}}. \bibinfo{pages}{257--265}.
\newblock


\bibitem[Chicano et~al\mbox{.}(2017)]%
        {chicano2017optimizing}
\bibfield{author}{\bibinfo{person}{Francisco Chicano}, \bibinfo{person}{Darrell Whitley}, \bibinfo{person}{Gabriela Ochoa}, {and} \bibinfo{person}{Renato Tin{\'o}s}.} \bibinfo{year}{2017}\natexlab{}.
\newblock \showarticletitle{Optimizing one million variable NK landscapes by hybridizing deterministic recombination and local search}. In \bibinfo{booktitle}{\emph{Proceedings of the genetic and evolutionary computation conference}}. \bibinfo{pages}{753--760}.
\newblock


\bibitem[Chicano et~al\mbox{.}(2024)]%
        {superMasksGray}
\bibfield{author}{\bibinfo{person}{Francisco Chicano}, \bibinfo{person}{Darrell Whitley}, \bibinfo{person}{Gabriela Ochoa}, {and} \bibinfo{person}{Renato Tin\'{o}s}.} \bibinfo{year}{2024}\natexlab{}.
\newblock \showarticletitle{Generalizing and Unifying Gray-Box Combinatorial Optimization Operators}. In \bibinfo{booktitle}{\emph{Parallel Problem Solving from Nature – PPSN XVIII: 18th International Conference, PPSN 2024, Hagenberg, Austria, September 14–18, 2024, Proceedings, Part I}} (Hagenberg, Austria). \bibinfo{publisher}{Springer-Verlag}, \bibinfo{pages}{52–67}.
\newblock
\showISBNx{978-3-031-70054-5}


\bibitem[Chicano et~al\mbox{.}(2014)]%
        {chicano2014efficient}
\bibfield{author}{\bibinfo{person}{Francisco Chicano}, \bibinfo{person}{Darrell Whitley}, {and} \bibinfo{person}{Andrew~M Sutton}.} \bibinfo{year}{2014}\natexlab{}.
\newblock \showarticletitle{Efficient identification of improving moves in a ball for pseudo-boolean problems}. In \bibinfo{booktitle}{\emph{Proceedings of the 2014 annual conference on genetic and evolutionary computation}}. \bibinfo{pages}{437--444}.
\newblock


\bibitem[Deb and Goldberg(1993)]%
        {decFunc}
\bibfield{author}{\bibinfo{person}{Kalyanmoy Deb} {and} \bibinfo{person}{David~E. Goldberg}.} \bibinfo{year}{1993}\natexlab{}.
\newblock \showarticletitle{Sufficient Conditions for Deceptive and Easy Binary Functions}.
\newblock \bibinfo{journal}{\emph{Ann. Math. Artif. Intell.}} \bibinfo{volume}{10}, \bibinfo{number}{4} (\bibinfo{year}{1993}), \bibinfo{pages}{385--408}.
\newblock


\bibitem[Deb et~al\mbox{.}(1993)]%
        {decBimodalOld}
\bibfield{author}{\bibinfo{person}{Kalyanmoy Deb}, \bibinfo{person}{Jeffrey Horn}, {and} \bibinfo{person}{David~E. Goldberg}.} \bibinfo{year}{1993}\natexlab{}.
\newblock \showarticletitle{Multimodal Deceptive Functions}.
\newblock \bibinfo{journal}{\emph{Complex Systems}} \bibinfo{volume}{7}, \bibinfo{number}{2} (\bibinfo{year}{1993}).
\newblock


\bibitem[Dushatskiy et~al\mbox{.}(2021)]%
        {ellGomea}
\bibfield{author}{\bibinfo{person}{Arkadiy Dushatskiy}, \bibinfo{person}{Tanja Alderliesten}, {and} \bibinfo{person}{Peter A.~N. Bosman}.} \bibinfo{year}{2021}\natexlab{}.
\newblock \showarticletitle{A Novel Approach to Designing Surrogate-assisted Genetic Algorithms by Combining Efficient Learning of Walsh Coefficients and Dependencies}.
\newblock \bibinfo{journal}{\emph{ACM Trans. Evol. Learn. Optim.}} \bibinfo{volume}{1}, \bibinfo{number}{2}, Article \bibinfo{articleno}{5} (\bibinfo{date}{July} \bibinfo{year}{2021}), \bibinfo{numpages}{23}~pages.
\newblock
\href{https://doi.org/10.1145/3453141}{doi:\nolinkurl{10.1145/3453141}}


\bibitem[Goldman and Punch(2014)]%
        {P3Original}
\bibfield{author}{\bibinfo{person}{Brian~W. Goldman} {and} \bibinfo{person}{William~F. Punch}.} \bibinfo{year}{2014}\natexlab{}.
\newblock \showarticletitle{Parameter-less Population Pyramid}. In \bibinfo{booktitle}{\emph{Proceedings of the 2014 Annual Conference on Genetic and Evolutionary Computation}} (Vancouver, BC, Canada) \emph{(\bibinfo{series}{GECCO '14})}. \bibinfo{publisher}{ACM}, \bibinfo{pages}{785--792}.
\newblock


\bibitem[Hagberg and Conway(2020)]%
        {hagberg2020networkx}
\bibfield{author}{\bibinfo{person}{Aric Hagberg} {and} \bibinfo{person}{Drew Conway}.} \bibinfo{year}{2020}\natexlab{}.
\newblock \showarticletitle{Networkx: Network analysis with python}.
\newblock \bibinfo{journal}{\emph{URL: https://networkx. github. io}} (\bibinfo{year}{2020}).
\newblock


\bibitem[Heckendorn(2002)]%
        {heckendorn2002}
\bibfield{author}{\bibinfo{person}{R.~B. Heckendorn}.} \bibinfo{year}{2002}\natexlab{}.
\newblock \showarticletitle{Embedded Landscapes}.
\newblock \bibinfo{journal}{\emph{Evolutionary Computation}} \bibinfo{volume}{10}, \bibinfo{number}{4} (\bibinfo{year}{2002}), \bibinfo{pages}{345--369}.
\newblock


\bibitem[Herrmann et~al\mbox{.}(2016)]%
        {herrmann2016communities}
\bibfield{author}{\bibinfo{person}{Sebastian Herrmann}, \bibinfo{person}{Gabriela Ochoa}, {and} \bibinfo{person}{Franz Rothlauf}.} \bibinfo{year}{2016}\natexlab{}.
\newblock \showarticletitle{Communities of local optima as funnels in fitness landscapes}. In \bibinfo{booktitle}{\emph{Proceedings of the Genetic and Evolutionary Computation Conference 2016}}. \bibinfo{pages}{325--331}.
\newblock


\bibitem[Herrmann et~al\mbox{.}(2018)]%
        {herrmann2018pagerank}
\bibfield{author}{\bibinfo{person}{Sebastian Herrmann}, \bibinfo{person}{Gabriela Ochoa}, {and} \bibinfo{person}{Franz Rothlauf}.} \bibinfo{year}{2018}\natexlab{}.
\newblock \showarticletitle{PageRank centrality for performance prediction: the impact of the local optima network model}.
\newblock \bibinfo{journal}{\emph{Journal of Heuristics}}  \bibinfo{volume}{24} (\bibinfo{year}{2018}), \bibinfo{pages}{243--264}.
\newblock


\bibitem[Komarnicki et~al\mbox{.}(2024)]%
        {overlapsKommar}
\bibfield{author}{\bibinfo{person}{Marcin~Michal Komarnicki}, \bibinfo{person}{Michal~Witold Przewozniczek}, \bibinfo{person}{Renato Tin\'{o}s}, {and} \bibinfo{person}{Xiaodong Li}.} \bibinfo{year}{2024}\natexlab{}.
\newblock \showarticletitle{Overlapping Cooperative Co-Evolution for Overlapping Large-Scale Global Optimization Problems}. In \bibinfo{booktitle}{\emph{Proceedings of the Genetic and Evolutionary Computation Conference}} (Melbourne, VIC, Australia) \emph{(\bibinfo{series}{GECCO '24})}. \bibinfo{publisher}{Association for Computing Machinery}, \bibinfo{address}{New York, NY, USA}, \bibinfo{pages}{665–673}.
\newblock
\showISBNx{9798400704949}
\href{https://doi.org/10.1145/3638529.3654171}{doi:\nolinkurl{10.1145/3638529.3654171}}


\bibitem[Malan(2021)]%
        {malan2021survey}
\bibfield{author}{\bibinfo{person}{Katherine~Mary Malan}.} \bibinfo{year}{2021}\natexlab{}.
\newblock \showarticletitle{A survey of advances in landscape analysis for optimisation}.
\newblock \bibinfo{journal}{\emph{Algorithms}} \bibinfo{volume}{14}, \bibinfo{number}{2} (\bibinfo{year}{2021}), \bibinfo{pages}{40}.
\newblock


\bibitem[Mostert et~al\mbox{.}(2019)]%
        {mostert2019insights}
\bibfield{author}{\bibinfo{person}{Werner Mostert}, \bibinfo{person}{Katherine~M Malan}, \bibinfo{person}{Gabriela Ochoa}, {and} \bibinfo{person}{Andries~P Engelbrecht}.} \bibinfo{year}{2019}\natexlab{}.
\newblock \showarticletitle{Insights into the feature selection problem using local optima networks}. In \bibinfo{booktitle}{\emph{Evolutionary Computation in Combinatorial Optimization: 19th European Conference, EvoCOP 2019, Held as Part of EvoStar 2019, Leipzig, Germany, April 24--26, 2019, Proceedings 19}}. Springer, \bibinfo{pages}{147--162}.
\newblock


\bibitem[Munetomo and Goldberg(1999a)]%
        {linc}
\bibfield{author}{\bibinfo{person}{M. Munetomo} {and} \bibinfo{person}{D.E. Goldberg}.} \bibinfo{year}{1999}\natexlab{a}.
\newblock \showarticletitle{A genetic algorithm using linkage identification by nonlinearity check}. In \bibinfo{booktitle}{\emph{IEEE SMC'99 Conference Proceedings. 1999 IEEE International Conference on Systems, Man, and Cybernetics (Cat. No.99CH37028)}}, Vol.~\bibinfo{volume}{1}. \bibinfo{pages}{595--600 vol.1}.
\newblock
\href{https://doi.org/10.1109/ICSMC.1999.814159}{doi:\nolinkurl{10.1109/ICSMC.1999.814159}}


\bibitem[Munetomo and Goldberg(1999b)]%
        {GoldMonotonicity}
\bibfield{author}{\bibinfo{person}{Masaharu Munetomo} {and} \bibinfo{person}{David~E. Goldberg}.} \bibinfo{year}{1999}\natexlab{b}.
\newblock \showarticletitle{Linkage identification by non-monotonicity detection for overlapping functions}.
\newblock \bibinfo{journal}{\emph{Evol. Comput.}} \bibinfo{volume}{7}, \bibinfo{number}{4} (\bibinfo{date}{dec} \bibinfo{year}{1999}), \bibinfo{pages}{377–398}.
\newblock
\showISSN{1063-6560}
\href{https://doi.org/10.1162/evco.1999.7.4.377}{doi:\nolinkurl{10.1162/evco.1999.7.4.377}}


\bibitem[Ochoa and Chicano(2019)]%
        {ochoa2019local}
\bibfield{author}{\bibinfo{person}{Gabriela Ochoa} {and} \bibinfo{person}{Francisco Chicano}.} \bibinfo{year}{2019}\natexlab{}.
\newblock \showarticletitle{Local optima network analysis for MAX-SAT}. In \bibinfo{booktitle}{\emph{Proceedings of the Genetic and Evolutionary Computation Conference Companion}}. \bibinfo{pages}{1430--1437}.
\newblock


\bibitem[Ochoa et~al\mbox{.}(2008)]%
        {ochoa2008study}
\bibfield{author}{\bibinfo{person}{Gabriela Ochoa}, \bibinfo{person}{Marco Tomassini}, \bibinfo{person}{Seb{\'a}stien V{\'e}rel}, {and} \bibinfo{person}{Christian Darabos}.} \bibinfo{year}{2008}\natexlab{}.
\newblock \showarticletitle{A study of NK landscapes' basins and local optima networks}. In \bibinfo{booktitle}{\emph{Proceedings of the 10th annual conference on Genetic and evolutionary computation}}. \bibinfo{pages}{555--562}.
\newblock


\bibitem[Ochoa et~al\mbox{.}(2017)]%
        {ochoa2017understanding}
\bibfield{author}{\bibinfo{person}{Gabriela Ochoa}, \bibinfo{person}{Nadarajen Veerapen}, \bibinfo{person}{Fabio Daolio}, {and} \bibinfo{person}{Marco Tomassini}.} \bibinfo{year}{2017}\natexlab{}.
\newblock \showarticletitle{Understanding phase transitions with local optima networks: number partitioning as a case study}. In \bibinfo{booktitle}{\emph{Evolutionary Computation in Combinatorial Optimization: 17th European Conference, EvoCOP 2017, Amsterdam, The Netherlands, April 19-21, 2017, Proceedings 17}}. Springer, \bibinfo{pages}{233--248}.
\newblock


\bibitem[Pedregosa et~al\mbox{.}(2011)]%
        {pedregosa2011scikit}
\bibfield{author}{\bibinfo{person}{Fabian Pedregosa}, \bibinfo{person}{Ga{\"e}l Varoquaux}, \bibinfo{person}{Alexandre Gramfort}, \bibinfo{person}{Vincent Michel}, \bibinfo{person}{Bertrand Thirion}, \bibinfo{person}{Olivier Grisel}, \bibinfo{person}{Mathieu Blondel}, \bibinfo{person}{Peter Prettenhofer}, \bibinfo{person}{Ron Weiss}, \bibinfo{person}{Vincent Dubourg}, {et~al\mbox{.}}} \bibinfo{year}{2011}\natexlab{}.
\newblock \showarticletitle{Scikit-learn: Machine learning in Python}.
\newblock \bibinfo{journal}{\emph{the Journal of machine Learning research}}  \bibinfo{volume}{12} (\bibinfo{year}{2011}), \bibinfo{pages}{2825--2830}.
\newblock


\bibitem[Przewozniczek et~al\mbox{.}(2022)]%
        {dgga}
\bibfield{author}{\bibinfo{person}{Michal~W. Przewozniczek}, \bibinfo{person}{Renato Tin\'{o}s}, \bibinfo{person}{Bartosz Frej}, {and} \bibinfo{person}{Marcin~M. Komarnicki}.} \bibinfo{year}{2022}\natexlab{}.
\newblock \showarticletitle{On Turning Black - into Dark Gray-Optimization with the Direct Empirical Linkage Discovery and Partition Crossover}. In \bibinfo{booktitle}{\emph{Proceedings of the Genetic and Evolutionary Computation Conference}} (Boston, Massachusetts) \emph{(\bibinfo{series}{GECCO '22})}. \bibinfo{publisher}{Association for Computing Machinery}, \bibinfo{address}{New York, NY, USA}, \bibinfo{pages}{269–277}.
\newblock
\showISBNx{9781450392372}
\href{https://doi.org/10.1145/3512290.3528734}{doi:\nolinkurl{10.1145/3512290.3528734}}


\bibitem[Przewozniczek et~al\mbox{.}(2023)]%
        {FIHCwLL}
\bibfield{author}{\bibinfo{person}{Michal~Witold Przewozniczek}, \bibinfo{person}{Renato Tin\'{o}s}, {and} \bibinfo{person}{Marcin~Michal Komarnicki}.} \bibinfo{year}{2023}\natexlab{}.
\newblock \showarticletitle{First Improvement Hill Climber with Linkage Learning -- on Introducing Dark Gray-Box Optimization into Statistical Linkage Learning Genetic Algorithms}. In \bibinfo{booktitle}{\emph{Proceedings of the Genetic and Evolutionary Computation Conference}} (Lisbon, Portugal) \emph{(\bibinfo{series}{GECCO '23})}. \bibinfo{publisher}{ACM}, \bibinfo{pages}{946–954}.
\newblock
\showISBNx{9798400701191}


\bibitem[{Sun} et~al\mbox{.}(2019)]%
        {rdg3}
\bibfield{author}{\bibinfo{person}{Y. {Sun}}, \bibinfo{person}{X. {Li}}, \bibinfo{person}{A. {Ernst}}, {and} \bibinfo{person}{M.~N. {Omidvar}}.} \bibinfo{year}{2019}\natexlab{}.
\newblock \showarticletitle{Decomposition for Large-scale Optimization Problems with Overlapping Components}. In \bibinfo{booktitle}{\emph{Proc. IEEE Congr. Evol. Comput. (CEC)}}. \bibinfo{pages}{326--333}.
\newblock


\bibitem[Tari et~al\mbox{.}(2023)]%
        {tari2023global}
\bibfield{author}{\bibinfo{person}{Sara Tari}, \bibinfo{person}{Gabriela Ochoa}, \bibinfo{person}{Matthieu Basseur}, {and} \bibinfo{person}{S{\'e}bastien Verel}.} \bibinfo{year}{2023}\natexlab{}.
\newblock \showarticletitle{On the Global Structure of PUBOi Fitness Landscapes}. In \bibinfo{booktitle}{\emph{Proceedings of the Companion Conference on Genetic and Evolutionary Computation}}. \bibinfo{pages}{247--250}.
\newblock


\bibitem[Teixeira and Pappa(2022)]%
        {teixeira2022understanding}
\bibfield{author}{\bibinfo{person}{Matheus~C Teixeira} {and} \bibinfo{person}{Gisele~L Pappa}.} \bibinfo{year}{2022}\natexlab{}.
\newblock \showarticletitle{Understanding AutoML search spaces with local optima networks}. In \bibinfo{booktitle}{\emph{Proceedings of the Genetic and Evolutionary Computation Conference}}. \bibinfo{pages}{449--457}.
\newblock


\bibitem[Thierens and Bosman(2013)]%
        {ltga}
\bibfield{author}{\bibinfo{person}{Dirk Thierens} {and} \bibinfo{person}{Peter~A.N. Bosman}.} \bibinfo{year}{2013}\natexlab{}.
\newblock \showarticletitle{Hierarchical Problem Solving with the Linkage Tree Genetic Algorithm}. In \bibinfo{booktitle}{\emph{Proceedings of the 15th Annual Conference on Genetic and Evolutionary Computation}} \emph{(\bibinfo{series}{GECCO '13})}. \bibinfo{publisher}{ACM}, \bibinfo{pages}{877--884}.
\newblock


\bibitem[Thomson et~al\mbox{.}(2017)]%
        {thomson2017comparing}
\bibfield{author}{\bibinfo{person}{Sarah~L Thomson}, \bibinfo{person}{Fabio Daolio}, {and} \bibinfo{person}{Gabriela Ochoa}.} \bibinfo{year}{2017}\natexlab{}.
\newblock \showarticletitle{Comparing communities of optima with funnels in combinatorial fitness landscapes}. In \bibinfo{booktitle}{\emph{Proceedings of the Genetic and Evolutionary Computation Conference}}. \bibinfo{pages}{377--384}.
\newblock


\bibitem[Thomson et~al\mbox{.}(2022)]%
        {thomson2022fractal}
\bibfield{author}{\bibinfo{person}{Sarah~L Thomson}, \bibinfo{person}{Gabriela Ochoa}, {and} \bibinfo{person}{S{\'e}bastien Verel}.} \bibinfo{year}{2022}\natexlab{}.
\newblock \showarticletitle{The fractal geometry of fitness landscapes at the local optima level}.
\newblock \bibinfo{journal}{\emph{Natural Computing}} \bibinfo{volume}{21}, \bibinfo{number}{2} (\bibinfo{year}{2022}), \bibinfo{pages}{317--333}.
\newblock


\bibitem[Thomson et~al\mbox{.}(2023)]%
        {thomson2023randomness}
\bibfield{author}{\bibinfo{person}{Sarah~L Thomson}, \bibinfo{person}{Nadarajen Veerapen}, \bibinfo{person}{Gabriela Ochoa}, {and} \bibinfo{person}{Daan van~den Berg}.} \bibinfo{year}{2023}\natexlab{}.
\newblock \showarticletitle{Randomness in local optima network sampling}. In \bibinfo{booktitle}{\emph{Proceedings of the Companion Conference on Genetic and Evolutionary Computation}}. \bibinfo{pages}{2099--2107}.
\newblock


\bibitem[Tin\'{o}s et~al\mbox{.}(2022)]%
        {ilsDLED}
\bibfield{author}{\bibinfo{person}{Renato Tin\'{o}s}, \bibinfo{person}{Michal~W. Przewozniczek}, {and} \bibinfo{person}{Darrell Whitley}.} \bibinfo{year}{2022}\natexlab{}.
\newblock \showarticletitle{Iterated Local Search with Perturbation Based on Variables Interaction for Pseudo-Boolean Optimization}. In \bibinfo{booktitle}{\emph{Proceedings of the Genetic and Evolutionary Computation Conference}} (Boston, Massachusetts) \emph{(\bibinfo{series}{GECCO '22})}. \bibinfo{publisher}{ACM}, \bibinfo{pages}{296–304}.
\newblock
\showISBNx{9781450392372}


\bibitem[Tin{\'o}s et~al\mbox{.}(2022)]%
        {tinos2022iterated}
\bibfield{author}{\bibinfo{person}{Renato Tin{\'o}s}, \bibinfo{person}{Michal~W Przewozniczek}, {and} \bibinfo{person}{Darrell Whitley}.} \bibinfo{year}{2022}\natexlab{}.
\newblock \showarticletitle{Iterated local search with perturbation based on variables interaction for pseudo-boolean optimization}. In \bibinfo{booktitle}{\emph{Proceedings of the Genetic and Evolutionary Computation Conference}}. \bibinfo{pages}{296--304}.
\newblock


\bibitem[Tin\'{o}s et~al\mbox{.}(2015)]%
        {pxForBinary}
\bibfield{author}{\bibinfo{person}{Renato Tin\'{o}s}, \bibinfo{person}{Darrell Whitley}, {and} \bibinfo{person}{Francisco Chicano}.} \bibinfo{year}{2015}\natexlab{}.
\newblock \showarticletitle{Partition Crossover for Pseudo-Boolean Optimization}. In \bibinfo{booktitle}{\emph{Proceedings of the 2015 ACM Conference on Foundations of Genetic Algorithms XIII}} (Aberystwyth, United Kingdom) \emph{(\bibinfo{series}{FOGA '15})}. \bibinfo{publisher}{Association for Computing Machinery}, \bibinfo{address}{New York, NY, USA}, \bibinfo{pages}{137–149}.
\newblock
\showISBNx{9781450334341}
\href{https://doi.org/10.1145/2725494.2725497}{doi:\nolinkurl{10.1145/2725494.2725497}}


\bibitem[Treimun-Costa et~al\mbox{.}(2020)]%
        {treimun2020modelling}
\bibfield{author}{\bibinfo{person}{German Treimun-Costa}, \bibinfo{person}{Elizabeth Montero}, \bibinfo{person}{Gabriela Ochoa}, {and} \bibinfo{person}{Nicol{\'a}s Rojas-Morales}.} \bibinfo{year}{2020}\natexlab{}.
\newblock \showarticletitle{Modelling parameter configuration spaces with local optima networks}. In \bibinfo{booktitle}{\emph{Proceedings of the 2020 Genetic and Evolutionary Computation Conference}}. \bibinfo{pages}{751--759}.
\newblock


\bibitem[Verel et~al\mbox{.}(2012)]%
        {verel2012local}
\bibfield{author}{\bibinfo{person}{S{\'e}bastien Verel}, \bibinfo{person}{Fabio Daolio}, \bibinfo{person}{Gabriela Ochoa}, {and} \bibinfo{person}{Marco Tomassini}.} \bibinfo{year}{2012}\natexlab{}.
\newblock \showarticletitle{Local optima networks with escape edges}. In \bibinfo{booktitle}{\emph{Artificial Evolution: 10th International Conference, Evolution Artificielle, EA 2011, Angers, France, October 24-26, 2011, Revised Selected Papers 10}}. Springer, \bibinfo{pages}{49--60}.
\newblock


\bibitem[Virtanen et~al\mbox{.}(2020)]%
        {virtanen2020scipy}
\bibfield{author}{\bibinfo{person}{Pauli Virtanen}, \bibinfo{person}{Ralf Gommers}, \bibinfo{person}{Travis~E Oliphant}, \bibinfo{person}{Matt Haberland}, \bibinfo{person}{Tyler Reddy}, \bibinfo{person}{David Cournapeau}, \bibinfo{person}{Evgeni Burovski}, \bibinfo{person}{Pearu Peterson}, \bibinfo{person}{Warren Weckesser}, \bibinfo{person}{Jonathan Bright}, {et~al\mbox{.}}} \bibinfo{year}{2020}\natexlab{}.
\newblock \showarticletitle{SciPy 1.0: fundamental algorithms for scientific computing in Python}.
\newblock \bibinfo{journal}{\emph{Nature methods}} \bibinfo{volume}{17}, \bibinfo{number}{3} (\bibinfo{year}{2020}), \bibinfo{pages}{261--272}.
\newblock


\bibitem[Whitley(2019)]%
        {whitleyNext}
\bibfield{author}{\bibinfo{person}{D. Whitley}.} \bibinfo{year}{2019}\natexlab{}.
\newblock \showarticletitle{Next generation genetic algorithms: a user’s guide and tutorial}.
\newblock In \bibinfo{booktitle}{\emph{Handbook of Metaheuristics}}. \bibinfo{publisher}{Springer}, \bibinfo{pages}{245--274}.
\newblock


\bibitem[Whitley et~al\mbox{.}(2020)]%
        {transTokBounded}
\bibfield{author}{\bibinfo{person}{Darrell Whitley}, \bibinfo{person}{Hernan Aguirre}, {and} \bibinfo{person}{Andrew Sutton}.} \bibinfo{year}{2020}\natexlab{}.
\newblock \showarticletitle{Understanding Transforms of Pseudo-Boolean Functions}. In \bibinfo{booktitle}{\emph{Proceedings of the 2020 Genetic and Evolutionary Computation Conference}} (Canc\'{u}n, Mexico) \emph{(\bibinfo{series}{GECCO '20})}. \bibinfo{publisher}{Association for Computing Machinery}, \bibinfo{address}{New York, NY, USA}, \bibinfo{pages}{760–768}.
\newblock
\showISBNx{9781450371285}


\bibitem[Whitley and Ochoa(2022)]%
        {whitley2022local}
\bibfield{author}{\bibinfo{person}{Darrell Whitley} {and} \bibinfo{person}{Gabriela Ochoa}.} \bibinfo{year}{2022}\natexlab{}.
\newblock \showarticletitle{Local optima organize into lattices under recombination: an example using the traveling salesman problem}. In \bibinfo{booktitle}{\emph{Proceedings of the Genetic and Evolutionary Computation Conference}}. \bibinfo{pages}{757--765}.
\newblock


\bibitem[Whitley et~al\mbox{.}(2016)]%
        {GrayBoxWhitley}
\bibfield{author}{\bibinfo{person}{L.~Darrell Whitley}, \bibinfo{person}{Francisco Chicano}, {and} \bibinfo{person}{Brian~W. Goldman}.} \bibinfo{year}{2016}\natexlab{}.
\newblock \showarticletitle{Gray Box Optimization for Mk Landscapes Nk Landscapes and Max-Ksat}.
\newblock \bibinfo{journal}{\emph{Evol. Comput.}} \bibinfo{volume}{24}, \bibinfo{number}{3} (\bibinfo{date}{Sept.} \bibinfo{year}{2016}), \bibinfo{pages}{491–519}.
\newblock
\showISSN{1063-6560}
\href{https://doi.org/10.1162/EVCO_a_00184}{doi:\nolinkurl{10.1162/EVCO_a_00184}}


\bibitem[Yu et~al\mbox{.}(2005)]%
        {cyclicTrap}
\bibfield{author}{\bibinfo{person}{Tian-Li Yu}, \bibinfo{person}{Kumara Sastry}, {and} \bibinfo{person}{David~E. Goldberg}.} \bibinfo{year}{2005}\natexlab{}.
\newblock \showarticletitle{Linkage Learning, Overlapping Building Blocks, and Systematic Strategy for Scalable Recombination}. In \bibinfo{booktitle}{\emph{Proceedings of the 7th Annual Conference on Genetic and Evolutionary Computation}} \emph{(\bibinfo{series}{GECCO '05})}. \bibinfo{publisher}{ACM}, \bibinfo{pages}{1217–1224}.
\newblock


\bibitem[Zhou et~al\mbox{.}(2024)]%
        {zhou2024evolutionary}
\bibfield{author}{\bibinfo{person}{Ryan Zhou}, \bibinfo{person}{Jaume Bacardit}, \bibinfo{person}{Alexander~EI Brownlee}, \bibinfo{person}{Stefano Cagnoni}, \bibinfo{person}{Martin Fyvie}, \bibinfo{person}{Giovanni Iacca}, \bibinfo{person}{John McCall}, \bibinfo{person}{Niki van Stein}, \bibinfo{person}{David~J Walker}, {and} \bibinfo{person}{Ting Hu}.} \bibinfo{year}{2024}\natexlab{}.
\newblock \showarticletitle{Evolutionary Computation and Explainable AI: A Roadmap to Understandable Intelligent Systems}.
\newblock \bibinfo{journal}{\emph{IEEE Transactions on Evolutionary Computation}} (\bibinfo{year}{2024}).
\newblock


\end{thebibliography}
	
\end{document}